%% file: iclr2027_conference.tex
\documentclass{article} 
\usepackage{iclr2027_conference,times}

\input{math_commands.tex}

\usepackage{hyperref}
\usepackage{url}

\usepackage{graphicx}
\usepackage{fontawesome5}
\usepackage{booktabs}
\usepackage{tcolorbox}
\usepackage{makecell}
\usepackage[linesnumbered,ruled]{algorithm2e}
\usepackage{colortbl}
\usepackage{float}
\usepackage{placeins}
\usepackage{stfloats}
\usepackage{enumitem}
\usepackage{multirow}
\usepackage{threeparttable}
\usepackage{amssymb}
\usepackage{adjustbox}
\usepackage{lineno}
\usepackage[x11names]{xcolor}
\usepackage{caption}
\usepackage{amsmath}
\usepackage{amsthm}
\usepackage{xspace}
\providecommand{\RAAGcleancopy}{}
\input{revision_marks}

\newcommand{\basehl}{\rowcolor{black!7}}
\newcommand{\ourhl}{\rowcolor{green!14}}     
\newcommand{\wamhl}{\rowcolor{red!12}}       
\newcommand{\retrhl}{\rowcolor{blue!9}}      

\newcommand{\method}{\textsc{RAAG}\xspace}

\title{RAAG: Training-Free Retrieval-Augmented Action Generation Guided by Gripper--Object Geometry from Successful Experience}

\author{
\begin{tabular}[t]{@{}l@{}}
\textbf{Jiaqi Li}{\normalfont\textsuperscript{1}}  \quad
\textbf{Guangming Wang}{\normalfont\textsuperscript{2}}  \quad
\textbf{Shuntian Zheng}{\normalfont\textsuperscript{1}}  \quad
\textbf{Zitong Zhang}{\normalfont\textsuperscript{1}}  \quad
\textbf{Hoang Hai Pham}{\normalfont\textsuperscript{1}}  \\
\textbf{Yu Fang}{\normalfont\textsuperscript{3}}  \quad
\textbf{Sai Gu}{\normalfont\textsuperscript{1}}  \quad
\textbf{Hesheng Wang}{\normalfont\textsuperscript{3}}  \quad
\textbf{Yu Guan}{\normalfont\textsuperscript{1\,\faEnvelope[regular]}}
\\[6pt]
\normalfont\small\textsuperscript{1}The University of Warwick  \quad
\textsuperscript{2}University of Cambridge \\
\normalfont\small\textsuperscript{3}Shanghai Jiao Tong University
\end{tabular}
}

\iclrfinalcopy
\begin{document}

\maketitle

\input{paper_sec/0_abstract}
\input{paper_sec/1_intro}
\input{paper_sec/2_related_work}
\input{paper_sec/3_method}
\input{paper_sec/4_exp}
\input{paper_sec/5_conclusion}

\subsection*{AI Use Statement}
In this work, we used generative AI tools to assist with language polishing, grammar and spelling correction, code editing, and numerical and implementation consistency checks with the manuscript.
We have not used generative AI tools for other research tasks requiring disclosure.
We have reviewed all AI-assisted work. We take responsibility for the final content of this work, including text, claims or artifacts produced with the aid of generative AI.

\subsection*{Ethics statement}
This work studies robotic manipulation using existing benchmarks, simulated environments, and a real-robot setup and involves no human subjects or personal data. As with other robotic control methods, deployment on physical systems should include appropriate safety validation and human oversight.

\subsection*{Reproducibility Statement}
We provide the RAAG formulation and evaluation protocol in the paper.
The appendix details perception and implementation, memory construction, benchmark settings, hyperparameters, sensitivity analyses, failure cases, and cross-environment evaluation.
Upon publication, we will release the code and scripts, together with the complete real-robot training dataset, trained model checkpoints, robot model files, hardware and software configuration guides, and 3D-printable models of the manipulation objects.

\clearpage
\bibliographystyle{iclr2027_conference}
\bibliography{main,raag}

\clearpage
\appendix
\input{paper_sec/6_appx}
\begin{RAAGchangeblock}
\input{paper_sec/7_perception_execution}
\end{RAAGchangeblock}
\begin{RAAGchangeblock}
\input{paper_sec/8_controls}
\end{RAAGchangeblock}
\begin{RAAGchangeblock}
\input{paper_sec/9_calibration}
\end{RAAGchangeblock}

\end{document}

%% file: math_commands.tex
\usepackage{amsmath,amsfonts,bm}

\def\eqref#1{equation~\ref{#1}}

\def\1{\bm{1}}

\DeclareMathAlphabet{\mathsfit}{\encodingdefault}{\sfdefault}{m}{sl}
\SetMathAlphabet{\mathsfit}{bold}{\encodingdefault}{\sfdefault}{bx}{n}



%% file: revision_marks.tex
\usepackage{etoolbox}
\newif\ifRAAGchanges
\ifdefined\RAAGcleancopy
  \RAAGchangesfalse
\else
  \RAAGchangestrue
\fi
\newcommand{\RAAGchange}[1]{\ifRAAGchanges\textcolor{blue}{#1}\else#1\fi}
\newcommand{\RAAGchangecolor}{\ifRAAGchanges\color{blue}\arrayrulecolor{black}\captionsetup{font+={color=blue}}\fi}
\newenvironment{RAAGchangeblock}{\ifRAAGchanges\color{blue}\fi}{}
\newcommand{\RAAGmarkedbibitem}[2][]{%
  \color{black}%
  \ifRAAGchanges
    \ifboolexpr{
      test {\ifstrequal{#2}{ames2017control}} or
      test {\ifstrequal{#2}{dragan2013policy}} or
      test {\ifstrequal{#2}{johannink2019residual}} or
      test {\ifstrequal{#2}{yang2021fudge}}
    }{\color{blue}}{}%
  \fi
  \RAAGsavedbibitem[#1]{#2}}
\AtBeginEnvironment{thebibliography}{%
  \let\RAAGsavedbibitem\bibitem
  \let\bibitem\RAAGmarkedbibitem}

%% file: paper_sec/0_abstract.tex
\begin{abstract}
Vision-language-action (VLA) policies generate actions from current observations, which can be brittle under distribution shift due to the lack of reliable guidance about future interaction outcomes. 
World models provide such foresight by predicting future states, but add training or test-time cost.
Retrieval offers a training-free alternative, yet memory-based methods often reuse recorded actions tied to the original scene geometry, while geometry-aware alternatives typically require heavier 3D perception.
We observe that successful trajectories contain two distinct signals: a scene-specific motion path and the relative gripper--object geometry that defines successful interaction.
The path may become invalid after scene and position changes, while the relative geometry can remain transferable and serve as a future target even in unseen cases.
Based on this insight, we propose RAAG, a training-free framework that augments frozen VLAs with future targets retrieved from successful memories. 
RAAG retrieves geometrically compatible experience using gripper--object relations, retargets successful relations to the current scene, and gates guidance by policy risk and retrieval reliability. 
The frozen VLA still determines how to reach these targets, avoiding brittle action reuse.
Across five benchmarks, two simulators, and a real robot, RAAG substantially improves frozen VLAs under distribution shifts, raising OpenVLA-OFT's success rate from 24.0\% to 69.6\% on ManiSkill3 OOD.
These gains come with modest inference overhead of 38.6\% additional latency on a single GPU and only 3.0 ms on dual GPUs, where retrieval runs in parallel with VLA inference.
Code and data are available at \href{https://github.com/JiaqiLi404/RAAG}{\nolinkurl{github.com/JiaqiLi404/RAAG}}.
\end{abstract}

%% file: paper_sec/1_intro.tex
\section{Introduction}
Embodied AI aims to teach robots to robustly manipulate the physical world, widely applied to household assistance \citep{fu2024mobile}, warehouse logistics \citep{mahler2019learning}, and assistive healthcare \citep{losey2022learning}. 
Recent progress has been driven by vision-language-action (VLA) policies, which generate temporally extended action chunks from current observation and a language instruction \citep{black2024pi_0,kim2025openvlaoft}.
While efficient, this design is brittle because actions are generated without explicitly modeling physical consequences, making small pose or appearance shifts enough to collapse performance \citep{fei2026libero,liu2025eva}.

\begin{figure}[tb]
  \centering
  \includegraphics[width=\linewidth]{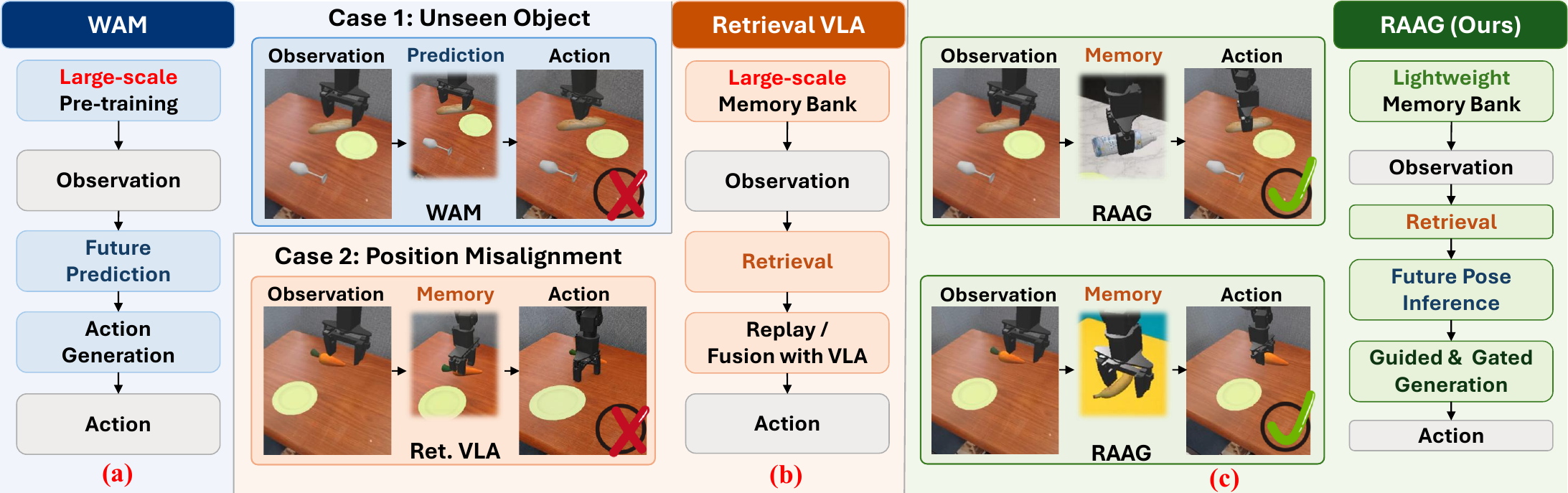}
\caption{
\textbf{Motivation of RAAG.}
(a) WAMs require large-scale pretraining, while learned world predictions can become unreliable under unseen scenes or objects, as shown in Case 1.
(b) Retrieval VLAs require large-scale memory banks and can fail under memory-scene misalignment in position or appearance, as shown in Case 2.
(c) Our RAAG extracts transferable gripper--object relations from successful past trajectories and uses them as future guidance, enabling a lightweight memory bank to generalize experience across similarly shaped object instances and unseen scenes.
}
\label{fig:motivation}
\end{figure}

World action models (WAMs) address this action-consequence link by making the future an intermediate variable for control.
They predict future observations in pixel \citep{ye2026world,cen2025worldvla,zhao2025cot,tian2025predictive} or latent-space \citep{zhang2026dreamvla,ye2025latent}, or learn future dynamics through large-scale pretraining \citep{sun2026vla}.
However, obtaining such future information requires additional learning or prediction, introducing extra expensive training or inference cost.
More importantly, as shown in Fig. \ref{fig:motivation}(a), errors in future prediction can directly propagate to action generation under distribution shift \citep{park2026test,zhao2025cot}.

In parallel, retrieval-based methods reuse successful past experience instead of explicitly predicting the future at test time
\citep{zhang2026retrieval,sridhar2025ricl,li2025map}.
Yet existing methods typically retrieve visually-matched experience and reuse scene-dependent actions through replay or fusion
\citep{kwon2025rt,zhao2026retrieve}.
As illustrated in Fig.~\ref{fig:motivation}(b), visual similarity does not guarantee geometric consistency, while pose shifts can make previously successful actions non-executable.
Memory can therefore become unreliable precisely when the test scene differs most from past experience, and can even degrade rollouts that the base policy would otherwise solve.

The key insight is that a successful trajectory contains two distinct kinds of information, \emph{what physical relation must be reached for success} and \emph{how that relation was reached in a particular scene}.
The latter is scene-specific, since the recorded actions depend on the original object and gripper poses and may become invalid after a scene shift.
In contrast, the gripper--object relation at a successful state captures the relative direction, distance, and orientation required for interaction, and thus can remain valid after the object moves.
More importantly, it comes from an actual successful execution rather than a learned prediction, making it an empirically supported future condition for the current rollout.
We therefore transfer this relation rather than the recorded trajectory, and let the frozen VLA determine how to reach it in the current scene.

Building on this idea, we propose \textbf{R}etrieval-\textbf{A}ugmented \textbf{A}ction \textbf{G}eneration (\method), a training-free framework that augments frozen VLAs with recalled future guidance.
Unlike WAMs that explicitly predict the future, and retrieval VLAs that reuse scene-dependent actions, \method retrieves verified gripper--object relations from past successes and uses them as future constraints for action generation.
Specifically, it retrieves physically compatible experience using gripper--object geometry and object shape, then retargets successful localization and interaction relations to guide the policy toward successful interaction.
The frozen VLA still generates the actions, while confidence gating controls how strongly memory intervenes based on policy risk and retrieval reliability.
We extensively evaluate \method across five benchmarks, two simulators, and a real robot, with a particular focus on memory transferability under distribution shifts.
By transferring only compact relational targets, \method avoids explicit future prediction and dense near-matched memory, improving success rate by up to $14.3\%$ on standard benchmark LIBERO-Plus \citep{fei2026libero} and $45.6\%$ on ManiSkill3 OOD \citep{tao2025maniskill3}. 
For discrete-token VLAs, \method adds 38.6\% inference latency on a single GPU and only 3.0 ms on dual GPUs when retrieval runs in parallel with VLA inference.


Our contributions are summarized as follows:
\begin{enumerate}[itemsep=0pt,topsep=0pt,parsep=0pt,leftmargin=10pt]
\item We identify the transferable part of memory. 
While recorded trajectories are tied to the past scene, successful gripper--object relations can remain valid after scene changes and serve as future targets.
This reframes retrieval from reusing past actions to transferring the future objective.

\item We propose \method, a training-free framework that realizes this idea for frozen VLAs.
\method retrieves physically compatible memories using gripper--object geometry, retargets it to the current scene, and gates guidance by policy risk and retrieval reliability.
It requires no learned future predictor and leaves action generation to the frozen policy.

\item  We extensively evaluate \method across five benchmarks, two simulators, and a real robot under distribution shifts.
\method improves OpenVLA-OFT's success rate from 24.0\% to 69.6\% on ManiSkill3, with only 3.0 ms added latency when perception model runs on an additional GPU.


\end{enumerate}


%% file: paper_sec/2_related_work.tex
\section{Related Work}

\textbf{Retrieval and test-time guidance.}
Retrieval methods reuse experience through replay \citep{pari2021surprising,kwon2025rt}, action fusion \citep{sridhar2025ricl}, or in-context conditioning \citep{zhang2026retrieval,li2025map}.
However, visual similarity does not guarantee geometric compatibility.
Retrieve-then-Steer \citep{zhao2026retrieve} uses confidence-adaptive action priors, while RA-VLA \citep{jang2026ra} learns behavior-aligned retrieval.
These improve memory integration, but object-pose changes can still misalign retrieved actions.
RAM \citep{kuang2024ram} addresses this mismatch through depth camera and explicit 3D affordance alignment with additional perception overhead.


Beyond retrieval, TACO \citep{yang2025steering} verifies action proposals, while ACG \citep{park2025acg} promotes temporal coherence.
However, coherent actions do not guarantee geometrically valid interactions.
AffordDP \citep{wu2025afforddp} incorporates transferable affordances into action generation, but requires training an affordance-conditioned diffusion policy.
\method instead retrieves gripper--object targets from sparse successful trajectories to guide frozen VLAs without additional training, allowing existing motion skills to pursue interaction goals adapted to the current scene.
This enables training-free transfer without closely matched memories.
Further comparisons are in Appx.~\ref{sec:appx:baselines}.

\textbf{World action models.}
In parallel, WAMs incorporate future information through visual prediction \citep{zhao2025cot,cen2025worldvla,ye2026world}, compact world representations \citep{zhang2026dreamvla}, or predictive pretraining \citep{sun2026vla,ye2025latent}.
These approaches provide learned priors for control but require large-scale predictive learning or additional future-generation computation.
Moreover, visually plausible predictions need not yield executable behavior \citep{jiang2026robowm}.
\method obtains future targets from the geometry of successful executions and retargets them to the current scene, providing guidance without training an additional future predictor.

%% file: paper_sec/3_method.tex
\section{Method}
\label{sec:method}
\subsection{Problem Definition}
We study language-conditioned robotic manipulation with a frozen VLA policy $\pi$ and a fixed memory bank $\mathcal{B}=\{\tau_i\}_{i=1}^{N}$ of successful rollouts.
Each trajectory records observations, robot states, and executed actions.
Given an instruction, the policy predicts an action chunk from the current RGB observation and proprioceptive state.
The robot executes the chunk and replans until success or budget exhaustion.
Our goal is to improve task success without additional training when changes in position, appearance, or instance make the current scene differ from the stored memory.

\textbf{RAAG Overview.}
RAAG uses successful gripper--object relations as future targets for action generation.
Our motivation is that appearance changes need not alter the interaction geometry, whereas position changes can invalidate a recorded action trajectory.
Transferring relative geometry therefore allows past successes to guide actions across these changes without requiring an identical memory.

\begin{figure}[H]
  \centering
  \includegraphics[width=0.96\linewidth]{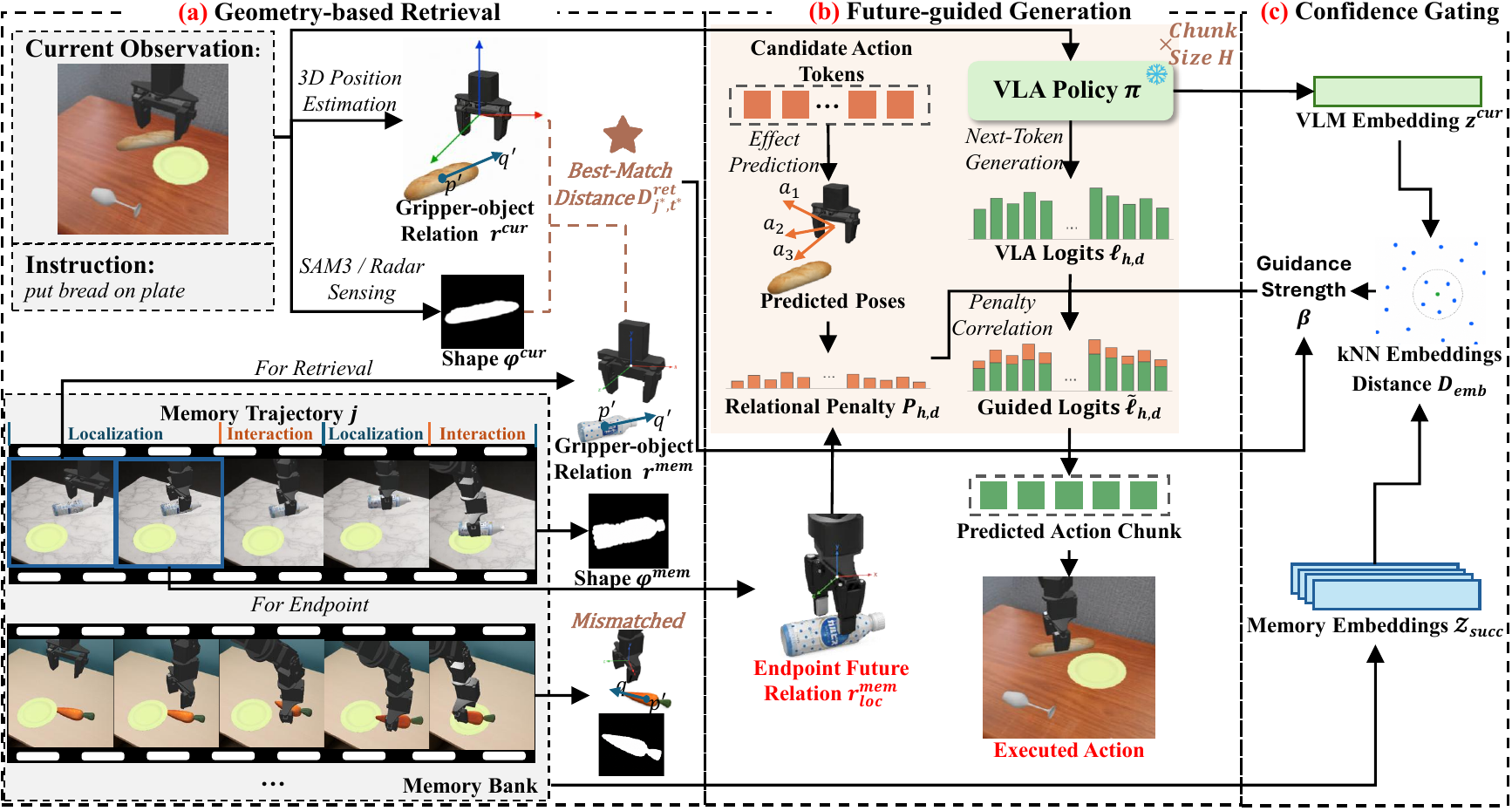}
\caption{
\textbf{Overview of our RAAG.}
(a) GR retrieves the best-matched successful trajectory using gripper--object geometry and object shape.
(b) FG splits execution into localization and interaction stages, uses each stage’s endpoint relation as the future target, and penalizes candidate action tokens by their predicted distance to that target.
Penalty computation runs in parallel with VLA inference.
(c) CG combines VLA reliability and retrieval confidence to adaptively adjust the guidance strength.
}
\label{fig:methodology}
\end{figure}

As shown in Fig.~\ref{fig:methodology}, RAAG consists of three complementary components.
\textit{Geometry-based Retrieval (GR)} matches gripper--object relations and object shape to retrieve geometrically compatible experience, reducing sensitivity to appearance and placement.
\textit{Future-guided Generation (FG)} uses retrieved endpoint relations to specify where the gripper should reach, while the frozen VLA generates how to reach it from the current state.
\textit{Confidence Gating (CG)} adjusts guidance strength using both estimated policy risk and retrieval reliability, strengthening intervention when the policy needs support and a compatible memory is available, while limiting unnecessary or unreliable guidance.

\subsection{Geometry-based Retrieval}
\label{sec:state}
GR aims to retrieve experience with compatible interaction geometry across changes in appearance and placement.
We therefore represent each state by the object pose relative to the gripper, complemented by object shape to account for differences in contact geometry.

\textbf{gripper--object relation.}
Interaction depends on the gripper's pose relative to the object, motivating a representation that separates interaction geometry from absolute scene placement.
Let $(p_t,q_t)$ and $(p',q')$ denote the world-frame poses of the gripper and the target object, respectively, with positions $p_t,p'\in\mathbb{R}^3$ and orientations $q_t,q'\in\mathbb{R}^3$.
Using $R(\cdot)$ to map an orientation to a rotation matrix and 
\vspace{-4pt}
$\operatorname{Log}(\cdot)$ to return its axis-angle vector, we express the object pose in the gripper frame as:
\begin{equation}
r =
\begin{bmatrix}
p^{\mathrm{rel}} \\
q^{\mathrm{rel}}
\end{bmatrix},
\qquad
p^{\mathrm{rel}} = R(q_t)^\top(p'-p_t),
\qquad
q^{\mathrm{rel}} =
\operatorname{Log}\!\left(R(q_t)^\top R(q')\right),
\label{eq:world_to_gripper}
\end{equation}
where $p^{\mathrm{rel}}$ and $q^{\mathrm{rel}}$ are the relative position and orientation, respectively.

The resulting relation $r$ supports both geometric cross-scene retrieval and endpoint transfer for action generation.
To obtain it from memory and online observations, we use robot kinematics for gripper pose and a replaceable perception module for object pose.
Our lightweight RGB-only implementation estimates object position and planar orientation from SAM~3 mask centroids and principal axes \citep{carion2026sam3} using standard camera calibration.
Details are provided in Appx.~\ref{app:relative_position}.

\textbf{Target shape.}
Similar relative poses can require different grasps for differently shaped objects.
We therefore complement relational matching with binary silhouettes, which retain object contours without texture or background appearance.
The current and memory silhouettes $\varphi^{cur}$ and $\varphi^{mem}$ are centered on a common grid using shared scaling that preserves aspect ratios.
Their mismatch is
\begin{equation}
D_{\mathrm{shape}}
=
1-\frac{|\varphi^{cur}\cap\varphi^{mem}|}
{|\varphi^{cur}\cup\varphi^{mem}|}.
\label{eq:shape_iou}
\end{equation}
Memory collection and standard evaluation share a fixed-camera setup, making silhouette overlap a useful, though viewpoint-dependent, cue to grasp compatibility.

\textbf{Retrieval.}
We retrieve experience compatible with the current interaction state: relative geometry provides a prior on approach direction and interaction stage, while shape constrains grasp compatibility.
Within the task-specific memory bank $\{\tau_j\}_{j=1}^{J}$ , indexed by the instruction’s action verb, we retrieve the trajectory and frame that minimize
\begin{equation}
D^{ret}_{j,t}
=
\left\|r^{cur}\ominus r_{j,t}^{mem}\right\|_2
+
w_s D_{\mathrm{shape},j,t},
\qquad
(j^\star,t^\star)
=
\arg\min_{j,t}D^{ret}_{j,t},
\label{eq:retrieval}
\end{equation}
where $\ominus$ measures the normalized geometric difference between the current relation $r^{cur}$ and each precomputed memory relation $r_{j,t}^{mem}$ (Appx.~\ref{app:relation_distance}), and $w_s$ weights the corresponding shape mismatch.
Minimizing this score identifies a matched frame $t^\star$ in trajectory $\tau_{j^\star}$, whose subsequent endpoint provides FG's future target.
For efficient retrieval, memory relations and shapes are precomputed.

Near contact, coarse relative position is insufficient because minor contact
differences can change the grasp action.
We therefore refine retrieval with interaction embeddings extracted by the frozen VLA visual encoder from background-masked crops of the gripper and target. 
Let $e^{cur}$ and $e_{j,t}^{mem}$ denote the embeddings of current observation and memory frame, the near-contact retrieval is operated by
\begin{equation}
D^{near}_{j,t}
=
D^{ret}_{j,t}
+
w_e\left\|e^{cur}-e_{j,t}^{mem}\right\|_2,
\qquad
(j^\star,t^\star)
=
\arg\min_{j,t}D^{near}_{j,t},
\label{eq:near_contact_retrieval}
\end{equation}
where $w_e$ weights visual evidence, complementing geometric compatibility with finer cues.

\subsection{Future-guided Generation}
\label{sec:connector}
FG uses retrieved endpoint geometry to guide motion in two stages.
The localization stage brings the gripper near the object, while the interaction refines its alignment to grasp or release.
Each stage uses a separate target, allowing coarse approach and precise interaction to be guided independently.

\textbf{Endpoint targets.}
Given the best-matched trajectory retrieved by GR, FG extracts stage-specific endpoints as future targets for the current state.
Gripper closing or opening marks the interaction endpoint $E_{\mathrm{int}}^{mem}$, with relation $r_{\mathrm{int}}^{mem}$ calculated from Eq.~\ref{eq:world_to_gripper}.
The localization endpoint $E_{\mathrm{loc}}^{mem}$ is the first pose along the approach satisfying $\|r_{j,t}^{mem}\ominus r_{\mathrm{int}}^{mem}\|_2<\delta$, using the normalized geometric residual from Eq.~\ref{eq:retrieval}.
It marks the transition from coarse approach to fine-grained interaction.

The endpoint relations $r^\star\in\{r_{\mathrm{loc}}^{mem},r_{\mathrm{int}}^{mem}\}$ specifies the desired object pose at each stage's end. 
Because it is defined in the gripper frame, $r^\star$ transfers across changes in absolute placement.

\textbf{Guided generation.}
FG incorporates the target gripper--object relation $r^\star$ into action generation through a geometric penalty that favors motion toward the target.
Unlike directly correcting through inverse kinematics, this penalty biases the VLA's predictions while retaining its learned preferences.

For discrete Cartesian actions, FG guides next step toward the target while accounting for motion already selected within the chunk.
To determine the remaining correction, we first recover the desired world-frame gripper pose from the current object pose and target relation $r^\star$.
At step $h$, we compute the offset from the predicted pre-action gripper pose to this target.
The resulting offset, expressed in the policy's action coordinates and clipped to its single-step displacement limits, defines the reference $u_h$.
We then favor candidates that follow this correction by assigning the penalty
\begin{equation}
P_{h,d}(a)=
\left(\frac{a-u_{h,d}}{\sigma_d}\right)^2,
\label{eq:action_penalty}
\end{equation}
where $a$ is a candidate action value in Cartesian dimension $d$, and $\sigma_d$ is the corresponding scale from the policy's action-normalization statistics.
We incorporate this penalty into the original VLA logits $\ell_{h,d}(a)$ to guide token selection:
\begin{equation}
\tilde{\ell}_{h,d}(a)
=
\ell_{h,d}(a)-\beta P_{h,d}(a),
\label{eq:guided_logit}
\end{equation}
where CG determines the guidance strength $\beta$.
After selecting all action tokens at step $h$, we update the predicted gripper pose and recompute $u_{h+1}$, keeping the next correction consistent with the selected action prefix.
This design also maintains inference efficiency, as only lightweight geometric updates are performed sequentially, without additional VLA forward passes.

RAAG primarily targets discrete-token VLAs, whose logits expose candidate action probabilities for direct logits-based guidance.
For other VLAs, we instead generate multiple candidate action chunks and rerank them using the same geometric guidance.
This extends FG beyond token-based policies, but incurs additional inference overhead.
Details are provided in Appx.~\ref{app:rerank}.

\textbf{Stage transitions.}
Within the verb-specific memory bank, the retrieved trajectory's gripper-closing and opening events determine the sequence and number of localization--interaction stages.
These gripper-state transitions serve as stage-completion signals and can be replaced with task-specific events for other interactions.
RAAG switches from localization to interaction when no gripper pose predicted along the generated chunk further approaches the target.
Executing the designated closing or opening command completes the interaction, after which RAAG advances to the next stage pair or terminates the sequence once all expected events are complete.

\textbf{Interaction refinement.}
Near contact, small alignment errors can change the required action, demanding finer visual evidence and more frequent feedback.
RAAG therefore augments geometric retrieval with local gripper--object visual embeddings (Eq.~\ref{eq:near_contact_retrieval}) to distinguish contact-sensitive configurations.
To respond promptly as these configurations change, it shortens the execution horizon to $H'$ and repeats retrieval at each action-generation round.
Together, finer matching and more frequent replanning enable timely correction during grasp or release.

\subsection{Confidence Gating}
\label{sec:gate}
CG determines the guidance strength $\beta_t$ in Eq.~\ref{eq:guided_logit}. 
Memory should intervene strongly only when the frozen VLA is outside its reliable regime and the retrieved relation is compatible with the current scene. 
CG therefore estimates policy confidence and retrieval confidence separately.

\textbf{Policy Confidence.}
Motivated by prior evidence that VLA internal features distinguish successful and failing executions \citep{gu2026safe}, we use the action-conditioning embedding to estimate policy risk. 
Taking OpenVLA-OFT \citep{kim2025openvlaoft} as an example, we obtain this embedding by mean-pooling the VLM's final hidden states before the action head, which directly condition action generation after perceiving the visual observation, instruction, and proprioceptive state.

During memory construction, RAAG stores these embeddings as
$\mathcal{Z}_{succ}=\{z_i^{mem}\}$. At test time, it measures the distance from
the current embedding $z^{cur}$ to its $k$ nearest successful anchors:
\begin{equation}
D_{emb}
=
\frac{1}{k}
\sum_{z_i^{mem}\in\mathrm{kNN}_{k}(z^{cur},\mathcal{Z}_{succ})}
\left\|
z^{cur}-z_i^{mem}
\right\|_2 .
\label{eq:vla_confidence}
\end{equation}
A larger distance indicates its predicted actions are more likely to fall outside the reliable behavior distribution.
To obtain a bounded policy confidence, we normalize this distance as
\begin{equation}
c^{risk}
=
\frac{D_{emb}}
{D_{emb}+\tau_{emb}},
\label{eq:policy_risk}
\end{equation}
where $\tau_{emb}$ is automatically calibrated as the 95th percentile of $kNN$ distances among successful memory states, excluding anchors from the same trajectory, and fixed within each dataset.

\textbf{Retrieval Confidence.}
A risky policy state does not guarantee that the retrieved memory provides a transferable target. 
Let $D_{j^\star,t^\star}$ be the retrieval distance computed by Eq.~\ref{eq:retrieval} during localization and Eq.~\ref{eq:near_contact_retrieval} during interaction. 
We similarly convert the distance into a bounded confidence score:
\begin{equation}
c^{ret}
=
\frac{\tau_{ret}}
{D_{j^\star,t^\star}+\tau_{ret}},
\label{eq:retrieval_confidence}
\end{equation}
where $\tau_{ret}$ is calibrated as the 95th percentile of leave-one-trajectory-out retrieval distances among memories. 
It is estimated separately for localization and interaction and fixed within each dataset.

Finally, CG sets the guidance strength in Eq.~\ref{eq:guided_logit} by scaling the maximum $\beta_{max}$ with confidences:
\begin{equation}
\beta
=
\beta_{max}\,
c^{risk}\,
c^{ret}.
\label{eq:confidence_gate}
\end{equation}
Guidance is strengthened when estimated policy risk and retrieval confidence are both high, and attenuated otherwise.
When $\beta=0$, the original action logits are preserved.


%% file: paper_sec/4_exp.tex
\section{Experiments}
\label{sec:exp}

\subsection{Experimental Setting}
\textbf{Benchmarks.}
We evaluate RAAG across five benchmarks, two simulators, and a real robot, with a focus on its memory generalization capabilities.
These include four LIBERO \citep{liu2023libero} suites, LIBERO-Plus \citep{fei2026libero} under seven perturbation factors, ManiSkill3 \citep{tao2025maniskill3}
\textsc{PutOnPlate25} with $13$ OOD factors, and our FailureBench with $856$ unseen failure states.
We further test real-robot deployment in Appx.~\ref{app:real_robot} and cross-environment transfer on SimplerEnv Bridge \citep{li2024simplerenv} in Appx.~\ref{app:simpler}.
Detailed protocols and benchmarks are provided in Appx.~\ref{app:benchmarks}.

\textbf{Memory bank.}
As discussed in Sec.~\ref{sec:appx:baselines}, existing retrieval methods often require large banks to find near-matched trajectories rather than transferring experience across scene changes.
Since prior memory banks are not released, we follow their collection protocol \citep{kwon2025rt} to collect our lightweight bank.
For each suite, every training-set instruction-object-scene combination contributes one successful trajectory to the memory bank, while positions and OOD factors vary only at test time.
For example, in ManiSkill3, $16$ objects across $16$ scenes yield $256$ cases and memory trajectories.
Exact memory-bank sizes for all benchmarks are reported in Appx.~\ref{app:memory_bank}.

\textbf{Metrics.}
We compare against World Action Models (WAM) and retrieval VLAs.
Following \cite{li2026global}, we report mean episode success rate and mean latency per VLA decision.

\DeclareRobustCommand{\legendbox}[2]{%
  \begingroup
  \setlength{\fboxsep}{0.4pt}%
  \smash{\raisebox{0.1ex}{\colorbox{#1}{\strut #2}}}%
  \endgroup
}
\begin{table}[!t]
\centering
\caption{
\textbf{LIBERO and LIBERO-Plus success rates} (\%).
The upper reports standalone policies and \legendbox{red!12}{WAMs}; the lower compares \legendbox{gray!15}{base policies} with add-on \legendbox{blue!9}{retrieval VLAs} and \legendbox{green!14}{our RAAG}.
\textit{\#Mem} denotes sum memory trajectories for all suites, and \textit{Lat.} denotes per-decision latency (ms).
Best per-backbone results are \textbf{bold}.
$^{*}$ marks training-based methods, whose \textit{\#Mem} counts training samples.
}
\label{tab:libero-main}
\setlength{\tabcolsep}{2.6pt}
\renewcommand{\arraystretch}{1.05}
\resizebox{0.99\textwidth}{!}{%
\begin{tabular}{@{}l ccccc ccccccc c cc@{}}
\toprule
& \multicolumn{5}{c}{\textbf{LIBERO}} & \multicolumn{8}{c}{\textbf{LIBERO-Plus}} & & \\
\cmidrule(lr){2-6}\cmidrule(lr){7-14}
\textbf{Method} & Sp & Ob & Go & L & \textbf{Avg} & Cam & Rob & Lang & Li & Bg & N & Lay & \textbf{All} & \textbf{\#Mem} & \textbf{Lat.} \\
\midrule
\;OpenVLA \citep{kim2024openvla} & 84.7 & 88.4 & 79.2 & 53.7 & 76.5 & 0.8 & 3.5 & 23.0 & 8.1 & 34.8 & 15.2 & 28.5 & 15.6 & $-$ & $-$ \\
\;$\pi_{0}$-FAST \citep{pertsch2025fast}   & 96.4 & 96.8 & 88.6 & 60.2 & 85.5 & 65.1 & 21.6 & 61.0 & 73.2 & 73.2 & 74.4 & 68.8 & 61.6 & $-$ & $-$ \\
\;NORA \citep{hung2025nora}          & 92.2 & 95.4 & 89.4 & 74.6 & 87.9 & 2.2 & 37.0 & 65.1 & 45.7 & 58.6 & 12.8 & 62.1 & 39.0 & $-$ & $-$ \\
\;GR00T-N1.5 \citep{bjorck2025gr00t} & 96.5 & 98.5 & 91.0 & 91.5 & 94.4& 46.8 & 52.6 & 82.6 & 90.0 & 87.5 & 71.5 & 67.2 & 69.5 & $-$ & $-$ \\
\;UniVLA \citep{bu2025univla}        & 96.5 & 96.8 & 95.6 & 92.0 & 95.2 & 1.8 & 46.2 & 69.6 & 69.0 & 81.0 & 21.2 & 31.9 & 43.9 & $-$ & $-$ \\
\;MemoryVLA \citep{shi2026memoryvla} & 98.4 & 98.4 & 96.4 & 93.4 & 96.6 & $-$ & $-$ & $-$ & $-$ & $-$ & $-$ & $-$ & $-$ & $-$ & $-$ \\

\wamhl \;CoT-VLA \citep{zhao2025cot}     & 87.5 & 91.6 & 87.6 & 69.0 & 83.9 & $-$ & $-$ & $-$ & $-$ & $-$ & $-$ & $-$ & $-$ & $-$ & 5953 \\
\wamhl \;WorldVLA \citep{cen2025worldvla}   & 87.6 & 96.2 & 83.4 & 60.0 & 81.8 & 0.1 & 27.9 & 41.6 & 43.7 & 17.1 & 10.9 & 38.0 & 25.0 & $-$ & $-$ \\
\wamhl \;DreamVLA \citep{zhang2026dreamvla} & 97.5 & 94.0 & 89.5 & 89.5 & 92.6 & 28.8 & 37.1 & 80.7 & 59.2 & 21.1 & 43.2 & 65.4 & 48.5 & $-$ & 91 \\
\wamhl \;VLA-JEPA \citep{sun2026vla}        & 96.2 & 99.6 & 97.2 & 95.8 & 97.2 & 63.3 & 67.1 & 85.4 & 95.6 & 93.6 & 66.3 & 85.1 & 79.5 & $-$ & 74 \\
\wamhl \;ForeAct \citep{zhang2026foreact}    & 97.3 & 99.8 & 97.3 & 95.4 & 97.5 & $-$ & $-$ & $-$ & $-$ & $-$ & $-$ & $-$ & $-$ & $-$ & $-$ \\

\midrule
\basehl \;$\pi_{0}$ \citep{black2024pi_0}   & 96.8 & 98.8 & 95.8 & 85.2 & 94.2 & 17.3 & 6.7 & 63.6 & 86.5 & 82.1 & 17.6 & 74.2 & 46.3 & $-$ & 191.3 \\
\retrhl \;\;\;$+$ Retrieval-VLA \citep{zhang2026retrieval}      & 97.4 & 98.8 & 96.3 & 89.5 & 95.5 & $-$ & $-$ & $-$ & $-$ & $-$ & $-$ & $-$ & $-$ & 800 & $-$ \\
\retrhl \;\;\;$+$ ECHO$^*$ \citep{hu2026echo}          & 98.3 & 98.8 & \textbf{98.6} & \textbf{93.5} & \textbf{97.3 }& $-$ & $-$ & $-$ & $-$ & $-$ & $-$ & $-$ & \textbf{56.5} & 4500 & $-$ \\
\ourhl \;\;\;$+$ \textbf{\method{} (ours, sampling-based guidance)}   & \textbf{98.9} &\textbf{ 99.2} & 97.9 & 93.2 & \textbf{97.3} & \textbf{25.0} & \textbf{19.6} & \textbf{67.3} & \textbf{87.5} & \textbf{83.3} & \textbf{37.5} &\textbf{ 83.9} & 55.1 & \textbf{598} & 413.5 \\

\midrule
\basehl \;$\pi_{0.5}$ \citep{intelligence2025pi_} & 98.8 & 98.2 & 98.0 & 92.4 & 96.9 & 75.4 & 77.5 & 85.6 & 96.9 & \textbf{94.6} & 89.7 & 85.7 & 85.7 & $-$ & 243.9 \\
\retrhl \;\;\;$+$ OptimusVLA$^*$ \citep{li2026global}         & \textbf{99.6} & \textbf{99.8} & 98.4 & 96.4 & 98.6 & 65.7 & 77.7 & 87.1 & 92.2 & 94.5 & 88.5 & 86.6 & 83.8 & 6500 & 182.7 \\
\ourhl \;\;\;$+$ \textbf{\method{} (ours, sampling-based guidance)}  & \textbf{99.6} & 99.2 & \textbf{99.8} & \textbf{98.8} & \textbf{99.3} & \textbf{79.5} & \textbf{90.6} & \textbf{87.3} & \textbf{97.3} & 94.3 & \textbf{94.6} & \textbf{98.2} & \textbf{91.3} & \textbf{598} & 577.8 \\

\midrule
\basehl \;OpenVLA-OFT (L1 head; \citealp{kim2025openvlaoft}) & 97.6 & 98.4 & 97.9 & 94.5 & 97.1 & 56.4 & 31.9 & 79.5 & 88.7 & 93.3 & 75.8 & 74.2 & 69.6 & $-$ & 143.4 \\
\retrhl \;\;\;$+$ ProbeAct \citep{zhang2026probeact}          & $-$ & $-$ & $-$ & $-$ & $-$ & 63.8 & 40.3 & 82.0 & \textbf{93.6} & 93.5 & 76.8 & 80.9 & 74.1 & $-$ & $-$ \\
\retrhl \;\;\;$+$ $\mu$VLA$^*$ \citep{cherepanov2026mu}      & 93.0 & \textbf{99.4} & 96.6 & 95.8 & 96.2 & $-$ & $-$ & $-$ & $-$ & $-$ & $-$ & $-$ & $-$ & 4500 & 150.4 \\
\basehl \;OpenVLA-OFT (Discrete head) & 93.3 & 94.1 & 95.9 & 92.4 & 93.9 & 53.2 & 29.2 & 75.4 & 84.9 & 87.8 & 73.0 & 71.7 & 66.2 & $-$ & 147.3 \\
\ourhl \;\;\;$+$ \textbf{\method{} (ours, logit-bias guidance)} & \textbf{99.0} & 99.0 & \textbf{98.0} & \textbf{98.0} & \textbf{98.5} & \textbf{66.2} & \textbf{50.9} & \textbf{86.8} & 90.0 & \textbf{93.6} & \textbf{90.7} & \textbf{92.1} & \textbf{80.5} & \textbf{598} & 151.5 \\
\bottomrule
\end{tabular}}
\end{table}

\textbf{Implementation.}
All experiments follow official protocols and settings of each benchmark, using one NVIDIA L40 for VLA and a second for SAM~3 \citep{carion2026sam3}.
Reproduction training uses an NVIDIA A100.
We detail the evaluated policies in Appx.~\ref{app:policies} and select hyperparameters by grid search on in-domain set.
For GR (Eq.~\ref{eq:retrieval} and \ref{eq:near_contact_retrieval}), we set $w_s{=}0.4$ and $w_e{=}0.8$.
For FG, we set $\delta{=}2$ and an execution horizon $H'{=}2$ for interaction stage.
For CG (Eq.~\ref{eq:vla_confidence} and \ref{eq:confidence_gate}), we set $K{=}2$ and $\beta_{\max}{=}10$.
The thresholds $\tau_{\mathrm{emb}}$ and $\tau_{\mathrm{ret}}$ are calibrated automatically during memory construction.

\begin{RAAGchangeblock}
\textbf{Perception source.}
The main simulation results use GT masks because off-the-shelf SAM~3 is less reliable on simulated objects.
We fine-tune SAM~3 on LIBERO training memories and evaluate it in Tab.~\ref{tab:sam3_tasks} and Appx.~\ref{app:sam3}; real-robot experiments use off-the-shelf SAM~3.
\end{RAAGchangeblock}

\subsection{Main Results}
\label{sec:main}

\textbf{LIBERO and LIBERO-Plus.}
As shown in Tab.~\ref{tab:libero-main}, RAAG consistently improves all three frozen backbones.
On LIBERO-Plus under seven perturbations, it improves OpenVLA-OFT from $66.2\%$ to $80.5\%$.
Even for the strong $\pi_{0.5}$, RAAG gains $13.1\%$ and $12.5\%$ under robot-state and layout shifts.
In contrast, lighting and background shifts largely preserve interaction geometry, leaving little room for improvement.
This demonstrates that recorded trajectories may no longer be executable under scene changes, whereas the geometric relation required for interaction can remain valid.

This difference is most evident against retrieval VLAs.
Existing methods reuse scene-dependent trajectories, which can improve clean performance yet become harmful under distribution shift.
OptimusVLA improves $\pi_{0.5}$ LIBERO performance to $98.6\%$ but reduces LIBERO-Plus to $83.8\%$, whereas RAAG improves both to $99.3\%$ and $91.3\%$.
This is largely because RAAG using memory not to imitate, but as a transferable future goal that guides the action generation for current scene.

RAAG also differs from WAMs, where robustness depends on the reliability of the predicted future itself.
This becomes a bottleneck under distribution shift, consistent with recent findings that errors
in visual foresight directly propagate to action generation~\citep{park2026test}. WorldVLA and DreamVLA reach only $25.0\%$ and $48.5\%$ on LIBERO-Plus, while VLA-JEPA reaches $79.5\%$.
In contrast, training-free RAAG reaches $91.3\%$ with $\pi_{0.5}$.
These results suggest that effective foresight requires remaining reliable under shift, rather than simply a stronger future predictor.

\textbf{ManiSkill3 and FailureBench.}
Tab.~\ref{tab:maniskill-main} highlights the importance of how sparse experience is transferred to a new scene.
With $256$ in-distribution memories, RT-Cache and VINN achieve only $2.3\%$ and $6.2\%$ OOD success, respectively, as replayed actions remain tied to the original scene geometry
and can become misaligned after scene shifts. Our RAM adaptation addresses this limitation by retargeting the retrieved geometry to the current scene and reaching the resulting pose through visual servoing.

\begin{table}[htbp]
\centering
\caption{
\textbf{ManiSkill3 and our FailureBench success rates} (\%).
ID and OOD is averaged across in-/out-of-distribution conditions.
RT-Cache and VINN use VLA visual features for retrieval and then replay, while RAM directly executes transferred 3D affordances; none invokes VLA generation.
}
\label{tab:maniskill-main}
\setlength{\tabcolsep}{4pt}
\renewcommand{\arraystretch}{1.1}
\resizebox{\textwidth}{!}{%
\begin{tabular}{@{}l ccccc ccccc cc@{}}
\toprule
& \multicolumn{5}{c}{\textbf{Clean starts} (official)}
  & \multicolumn{5}{c}{\textbf{From failure states} (\textsc{FailureBench})} & & \\
\cmidrule(lr){2-6}\cmidrule(lr){7-11}
\textbf{Method}
& ID & Vision & Semantic & Exec. & \textbf{OOD}
& ID & Position & Appear. & Adjacent & \textbf{OOD} & \textbf{\#Mem} & \textbf{Lat.} \\
\midrule
\;OpenVLA \citep{kim2024openvla} & 45.3 & 39.8 & 38.6 & 36.4 & 38.3 & 26.6 & 9.8 & 15.8 & 21.7 & 15.8 & $-$ & 292.7 \\
\;\;$+$ TT-VLA \citep{liu2026fly} & $-$ & 41.9 & 41.5 & 39.8 & 41.1 & $-$ & $-$ & $-$ & $-$ & $-$ & $-$ & $-$ \\
\;\;$+$ TraceVLA \citep{zheng2025tracevla} & $-$ & 59.5 & 45.1 & 26.9 & 43.9 & $-$ & $-$ & $-$ & $-$ & $-$ & $-$ & $-$ \\

\midrule
\basehl OpenVLA-OFT (Discrete head; \citealp{kim2025openvlaoft})& 35.9  & 31.2 & 27.6 & 13.3 & 24.0 & 18.8 & 7.1  & 12.2 & 19.9 & 13.1 & $-$ & 84.3 \\
\retrhl\;\;$+$ VINN \citep{pari2021surprising} & 17.2 & 8.1  & 6.9  & 3.6  & 6.2  & 10.9 & 3.4  & 5.5  & 12.0 & 7.2  & 256 & $-$ \\
\retrhl\;\;$+$ RT-Cache \citep{kwon2025rt}             & 3.1  & 2.8  & 2.5  & 1.6  & 2.3  & 9.4  & 5.5  & 10.1 & 12.2 & 9.1  & 256 & 31.2 \\
\retrhl\;\;$+$ RAM (w. our 3D perception; \citealp{kuang2024ram}) & 29.7 & 23.9  & 28.9 & 23.7 & 25.5 & 14.1 & 17.8 & 19.4 & 18.0 & 18.4 & 256 & 58.5 \\
\retrhl\;\;$+$ Robomonkey \citep{kwok2025robomonkey} & 46.9 & 45.2 & 41.2 & 33.1 & 39.8 & 28.1 & 13.4  & 25.4 & 29.9 & 22.6 & $-$ & 319.9 \\
\ourhl \;\;\textbf{$+$ \method{} (ours, logit-bias guidance)} & \textbf{87.5} & \textbf{82.4} & \textbf{81.8} & \textbf{44.5} & \textbf{69.6}
                       & \textbf{37.5} & \textbf{33.1} & \textbf{42.0} & \textbf{43.6} & \textbf{39.1} & 256 & 87.3 \\
\midrule
\basehl OpenVLA-OFT-GRPO (Discrete head)    & 93.8 & 75.6 & 55.3 & 41.7 & 57.5 & 62.5 & 44.8 & 46.8 & 52.7 & 48.2 & $-$ & 84.3 \\
\retrhl\;\;$+$ VINN \citep{pari2021surprising} & 17.2 & 6.9  & 6.6  & 2.1  & 5.2  & 6.2  & 2.8  & 4.2  & 6.7  & 4.5  & 256 & $-$ \\
\retrhl\;\;$+$ RT-Cache \citep{kwon2025rt}              & 6.3  & 1.2  & 3.4  & 1.6  & 2.1  & 12.5 & 3.6  & 9.0  & 10.1 & 7.3  & 256 & 31.2 \\
\retrhl\;\;$+$ RAM (w. our 3D perception; \citealp{kuang2024ram}) & 20.3 & 24.3 & 30.1 & 23.8 & 26.1 & 12.5 & 15.2 & 21.0  & 18.5& 17.9 & 256 & 58.5 \\
\retrhl\;\;$+$ Robomonkey \citep{kwok2025robomonkey} & 93.8 & 85.2 & 66.0 & 45.6 & 65.6 & 67.2 & 49.2 & 52.4 & 55.0 & 52.2 & $-$ & 319.9 \\
\ourhl \;\;\textbf{$+$ \method{} (ours, logit-bias guidance)} & \textbf{96.9} & \textbf{92.8} & \textbf{87.2} & \textbf{61.5} & \textbf{80.5}
                       & \textbf{75.0} & \textbf{59.1} & \textbf{64.9} & \textbf{67.9} & \textbf{63.8} & 256 & 87.3 \\
\midrule
\basehl OpenVLA-OFT-PPO (Discrete head)     & 96.9 & 89.1 & 69.4 & 69.3 & 75.9 & 84.4 & 65.3 & 63.8 & 68.2 & 66.0 & $-$ & 84.3 \\
\retrhl\;\;$+$ VINN \citep{pari2021surprising} & 1.6  & 0.0  & 0.9  & 1.0  & 0.7  & 7.8  & 2.2  & 4.5  & 8.5  & 5.2  & 256 & $-$ \\
\retrhl\;\;$+$ RT-Cache \citep{kwon2025rt}              & 3.1  & 2.2  & 2.5  & 1.0  & 1.9  & 9.4  & 2.9  & 9.6  & 11.8 & 7.9  & 256 & 31.2 \\
\retrhl\;\;$+$ RAM (w. our 3D perception; \citealp{kuang2024ram}) & 18.8 & 26.7 & 29.8 & 24.5 & 27.0 & 14.1 & 20.3  & 22.1 & 21.6 & 21.2 & 256 & 58.5 \\
\retrhl\;\;$+$ Robomonkey \citep{kwok2025robomonkey} & 95.3 & 91.2 & 74.2 & 77.9 & 81.1 & \textbf{89.1} & 67.6 & 68.6 & 73.0 & 69.9 & $-$ & 319.9 \\
\ourhl \;\;\textbf{$+$ \method{} (ours, logit-bias guidance)} & \textbf{98.4} & \textbf{95.9} & \textbf{89.1} & \textbf{84.9} & \textbf{90.0}
                       & 87.5 & \textbf{76.0} & \textbf{77.7} & \textbf{77.7} & \textbf{77.0} & 256 & 87.3 \\
\bottomrule
\end{tabular}}
\end{table}

This improves OOD success to $25.5\%$, but execution remains directly dependent on the accuracy of the estimated pose, making coarse perception errors consequential for control.

RAAG retains this geometric retargeting while using the retrieved endpoint as soft guidance for the frozen VLA.
The geometric prior specifies the desired interaction, while the VLA's observation-conditioned action preferences help determine how to reach it.
This combination raises OpenVLA-OFT's OOD success from $24.0\%$ to $69.6\%$ and consistently improves the stronger GRPO and PPO policies.
Together, these results support transferring sparse experience through scene-adapted geometric targets, with the VLA's learned action prior reducing sensitivity to potential geometric errors.

FailureBench further increases the memory-scene mismatch by starting from unseen off-trajectory failure states, with no recovery demonstrations in memory.
Here, action reuse becomes even less reliable, whereas RAAG can still use successful memories as transferable future targets, improving Base/GRPO/PPO from $13.1/48.2/66.0\%$ to $39.1/63.8/77.0\%$, respectively.
This shows that geometric memory remains useful even after the object is displaced, reoriented, or even toppled.

\textbf{Efficiency.}
RAAG obtains guidance from reusable geometric relations rather than learned future prediction, avoiding both future-generation costs and additional training.
On our dual-GPU system, perception and retrieval overlap with VLA inference, leaving only a small sequential guidance overhead.
The latency increases by $3.0$\,ms with one view ($84.3\rightarrow87.3$\,ms) and $4.2$\,ms with two views ($147.3\rightarrow151.5$\,ms).
For discrete-token policies, RAAG directly biases action probabilities without additional policy forward passes, while flow-based policies require sampling and reranking, resulting in greater overhead.
Appx.~\ref{app:efficiency} provides stage-wise timings and single-GPU analysis.

Relational transfer also reduces memory requirements.
The same interaction can guide actions across scene changes without densely covering each configuration.
RAAG only uses $598$ trajectories, compared with $800$ for Retrieval-VLA, $4{,}500$ for training methods, and $6{,}500$ for OptimusVLA.

\subsection{Ablations}
\label{sec:ablations}

\textbf{Components.}
According to Tab.~\ref{tab:ablation}(a), increasing control frequency alone provides limited improvement, while SAM~3 masking retains only the gripper and object but reduces performance to $8.8/10.1$, mainly because these masked observations differ from the policy's training inputs.

Within RAAG, replacing geometric retrieval with appearance matching lowers it to $25.5/46.2$, while retaining geometric retrieval but replacing FG with replay causes a larger drop to $16.3/21.7$.
These results suggest that geometric retrieval reduces memory--scene misalignment, while geometry-guided generation further compensates for the residual mismatch.
\begin{table}[htbp]
    \centering
    \caption{
        \textbf{Ablation of RAAG.}
        (a) Contributions of auxiliary modifications and RAAG components.
        (b) Geometric compatibility in retrieval.
        (c) Comparisons of memory use in replaying, servoing, and guiding VLA.
        FB denotes FailureBench; MS3 denotes the official ManiSkill3 OOD suite.
    }
    \vspace{-3px}
    \label{tab:ablation}

    \begingroup
    \scriptsize
    \setlength{\tabcolsep}{1pt}
    \renewcommand{\arraystretch}{0.93}
    \begin{minipage}[t]{0.28\textwidth}
        \centering
        \begin{tabular*}{\linewidth}{
            @{}l@{\extracolsep{\fill}}rr@{}
        }
            \toprule
            Method & FB & MS3 \\
            \midrule
            Frozen policy
                & 13.1 & 24.0 \\
            + High-freq.\ ($H=2$)
                & 18.6 & 32.1 \\
            + SAM 3 masking
                & 8.8 & 10.1 \\
            \midrule
            \textbf{Full RAAG}
                & \textbf{39.1} & \textbf{69.6} \\
            $-$ Geometric retrieval
                & 25.5 & 46.2 \\
            $-$ Future guidance
                & 16.3 & 21.7 \\
            $-$ Confidence gating
                & 38.0 & 71.0 \\
            \bottomrule
        \end{tabular*}
    \end{minipage}
    \hfill
    \begin{minipage}[t]{0.25\textwidth}
        \centering
        \begin{tabular*}{\linewidth}{
            @{}l@{\extracolsep{\fill}}rr@{}
        }
            \toprule
            Selection / key & FB & MS3 \\
            \midrule
            Fixed selection
                & 3.8 & 9.5 \\
            Random selection
                & 15.8 & 29.9 \\
            Appearance only
                & 25.5 & 46.2 \\
            \RAAGchange{Relational prototype}
                & \RAAGchange{27.8} & \RAAGchange{46.4} \\
            \RAAGchange{Object prototype}
                & \RAAGchange{35.4} & \RAAGchange{64.8} \\
            \midrule
            \textbf{Full geometric key}
                & \textbf{39.1} & \textbf{69.6} \\
            $-$ Relation
                & 25.9 & 47.1 \\
            $-$ Shape
                & 33.8 & 64.0 \\
            $-$ Appearance
                & 38.4 & 68.9 \\
            \bottomrule
        \end{tabular*}
    \end{minipage}
    \hfill
    \begin{minipage}[t]{0.45\textwidth}
        \centering
        \begin{tabular*}{\linewidth}{
            @{}l@{\extracolsep{\fill}}rr@{}
        }
            \toprule
            Memory use & FB & MS3 \\
            \midrule
            Action replay w/o geo.\ retrieval
                & 0.6 & 3.9 \\
            Action replay w.\ geo.\ retrieval
                & 16.3 & 21.7 \\
            Visual servoing
                & 18.4 & 25.5 \\
            \midrule
            \textbf{Both endpoints (ours)}
                & \textbf{39.1} & \textbf{69.6} \\
            Localization only
                & 31.2 & 53.5 \\
            Interaction only
                & 27.5 & 40.7 \\
            \midrule
            $\beta_{max}$= 5/\textbf{10}/20 (Eq.~\ref{eq:confidence_gate})
                & 35.1/\textbf{39.1}/37.7
                & 58.8/\textbf{69.6}/68.9 \\
            \bottomrule
        \end{tabular*}
    \end{minipage}
    \endgroup
\end{table}

Finally, confidence gating controls when memory should intervene.
Aggressive guidance can help weak policies but may override reliable predictions from stronger ones, motivating the competence-dependent analysis in Appx.~\ref{sec:embood}.

\textbf{Geometry key.}
In Tab.~\ref{tab:ablation}(b), random retrieval selects one trajectory from the verb-specific bank per episode and retains it throughout, reducing success to $15.8/29.9$.
Using a single fixed trajectory across all action verbs further lowers success to $3.8/9.5$.
These results indicate that task compatibility alone is insufficient, the effective memory reuse also requires matching the current interaction geometry.
For geometric retrieval, removing the gripper--object relation ($r$ in Eq.~\ref{eq:world_to_gripper}) reduces performance to near appearance-only levels, highlighting the importance of geometric consistency for cross-scene transfer.
Relative geometry alone, however, cannot account for shape-dependent grasp requirements.
Shape ($D_{\mathrm{shape}}$ in Eq.~\ref{eq:shape_iou}) therefore constrains contact compatibility, while appearance ($w_e$ in Eq.~\ref{eq:near_contact_retrieval}) further refines matching near contact.
Further sensitivity analysis is in Appx.~\ref{app:sensitivity}.

\textbf{Memory use.}
Tab.~\ref{tab:ablation}(c) compares guidance at different stages: each ``only'' variant applies RAAG to the named stage and leaves the other to the frozen VLA.
Localization-only guidance outperforms interaction-only, suggesting that localization needs more support under scene shifts, while grasping skills transfer more readily once localized correctly.
Guiding both stages further improves performance to $39.1/69.6$, demonstrating the additional benefit of refining interaction after localization.

Geometric retrieval improves action replay from $0.6/3.9$ to $16.3/21.7$, highlighting the importance of geometric consistency.
Yet its limited success suggests that matching alone cannot resolve residual scene--memory misalignment.
Visual servoing performs explicit 3D affordance warping equalled to RAM \citep{kuang2024ram} in Tab. \ref{tab:maniskill-main}, which is highly sensitive to quality of 3D localization.
RAAG combines the geometric targets with the learned action prior, achieving $39.1/69.6$.
The $\beta_{\max}$ sweep supports this balance: moderate guidance outperforms both weaker and stronger intervention.

\begin{RAAGchangeblock}
\begin{samepage}
\textbf{Additional controls.}
Fixed prototypes still trail Full GR, indicating that task/object compatibility alone cannot fully replace current-state retrieval (Tab.~\ref{tab:ablation}(b); Appx.~\ref{app:prototypes}).
With matched inputs and a frozen backbone, FG improves over fusion rules that can retain motion conflicting with the endpoint, supporting guidance during action selection (Appx.~\ref{app:fusion_controls}).
The gains also persist with fine-tuned SAM~3 masks (Tab.~\ref{tab:sam3_tasks}), supporting guidance under predicted segmentation (Appx.~\ref{app:sam3}).
\par
\end{samepage}

\begin{table}[t]
\RAAGchangecolor
\centering
\caption{\textbf{End-to-end success with predicted object masks} (\%). All rows use the same frozen OpenVLA-OFT checkpoint within each benchmark.}
\label{tab:sam3_tasks}
\small
\setlength{\tabcolsep}{5pt}
\begin{tabular}{@{}lrrr@{}}
\toprule
Benchmark & Frozen VLA & \makecell{RAAG +\\simulator masks} & \makecell{RAAG +\\adapted SAM~3} \\
\midrule
ManiSkill3 OOD & 24.0 & 69.6 & 68.9 \\
FailureBench & 13.1 & 39.1 & 38.9 \\
LIBERO & 93.9 & 98.5 & 97.4 \\
\bottomrule
\end{tabular}
\end{table}
\end{RAAGchangeblock}
\FloatBarrier
\subsection{Discussion}
RAAG is limited by its lightweight RGB-only perception, which relies on SAM~3 and camera calibration.
\begin{RAAGchangeblock}
Inaccurate camera extrinsics can distort the gripper--object geometry and misdirect guidance.
We mitigate this by aligning visible robot landmarks with their forward-kinematic positions to self-calibrate the camera pose (Appx.~\ref{app:self_calibration}).
\end{RAAGchangeblock}
Single-view localization further assumes a tabletop, introducing height bias that multi-view triangulation can reduce (Appx.~\ref{appx:failure}).
Future work could incorporate depth cameras, LiDARs, or mmWave radars to provide richer 3D geometry.
To further validate RAAG in practical settings, we evaluate cross-environment transfer in Appx.~\ref{app:simpler} and real-robot performance in Appx.~\ref{app:real_robot}.
Finally, unnecessary memory intervention can disrupt strong policies.
CG mitigates this risk, but more reliable policy-aware gating remains an important direction (Appx.~\ref{sec:embood}).

%% file: paper_sec/5_conclusion.tex
\section{Conclusion}
We introduce RAAG, a training-free framework that guides frozen VLAs with gripper--object relations retrieved from successful executions.
By separating reusable interaction geometry from scene-dependent actions, RAAG retargets past experience to the current scene without learning a future predictor or requiring dense memories.
Our results demonstrate substantially improved memory transfer across diverse appearance and pose shifts and unseen cases, highlighting the value of geometric compatibility over near-identical scene coverage.
These results demonstrate the potential of a small demonstration memory to provide transferable future guidance, paving the way for embodied agents that adapt to complex new environments by reusing past successes without retraining.

%% file: paper_sec/6_appx.tex
\section{Comparison with retrieval-based manipulation methods}
\label{sec:appx:baselines}

\begin{table}[htbp]
\centering
\tiny
\begin{adjustbox}{width=1\textwidth,center}
\begin{tabular}{lcccccc}
\toprule
Method & Retrieve key & Execution  & Gated & RGB-only  & Train-free & Memory quantity\\
\midrule
RAM \citep{kuang2024ram}
& \textbf{geometry}
& 3D transfer
& 
& 
& \checkmark 
& 51 traj./case
\\
DINOBot \citep{di2024dinobot}
& vision
& replay
& 
& 
&  \checkmark
& \textbf{1 traj./case}
\\
RT-cache \citep{kwon2025rt}
& vision
& replay
& 
& \checkmark
&  \checkmark 
& millions traj.
\\
RiCL  \citep{sridhar2025ricl}
& vision
& prompt + fusion
& 
& \checkmark
& 
& 20 traj./task
\\
MAP-VLA \citep{li2025map}
& vision + pose
& prompt
& 
& \checkmark
&
& 38 traj./case
\\
OptimusVLA \citep{li2026global}
& vision + pose
& fusion
&
& \checkmark
&
& 50 traj./case
\\
ECHO \citep{hu2026echo}
& vision + pose
& fusion
& \checkmark
& \checkmark
&
& not reported
\\
Retrieval-VLA \citep{zhang2026retrieval}
& vision + pose
& fusion
& 
& \checkmark
&  \checkmark
& 20 traj./case
\\
AffordDP \citep{wu2025afforddp}
& vision
& \textbf{guidance}
& 
& 
&  
& 20 traj./case
\\
\midrule
\textbf{\method (ours)}
& \textbf{geometry}
& \textbf{guidance}
& \checkmark
& \checkmark
& \checkmark 
& \textbf{1 traj./case}
\\
\bottomrule
\end{tabular}
\end{adjustbox}
\caption{
\textbf{Comparison with retrieval-based manipulation methods.}
\textit{Retrieve key} specifies the representation used to match the current state with memory. 
\textit{Execution} indicates whether the retrieved experience is replayed, provided as a prompt, fused into the policy, or used as guidance.
\textit{RGB-only} denotes methods that avoid costly 3D affordance pipelines, and \textit{Gated} marks methods that explicitly suppress unreliable memory intervention.
\textit{Memory quantity} reports the number of trajectories collected per comparable task-object-scene case, where test-time changes mainly involve object position or unseen factors.
}
\label{tab:appx:related_work_retrieval}
\end{table}

The compared methods differ substantially in memory scale because most rely on dense coverage to find experience that closely matches the test scene.
Retrieval-VLA uses $20$ demonstrations per task, MAP-VLA distills approximately $378$ demonstrations into trained prompts, OptimusVLA trains on $6.5$K in-domain trajectories, and RT-Cache uses a million-scale bank.
This scale partly compensates for the limited transferability of replayed or fused, scene-specific experience. 

A small bank alone, however, does not guarantee transferable retrieval.
DINOBot \citep{di2024dinobot} uses one trajectory per case, but requires RGB-D perception and explicit trajectory warping.
In contrast, \method combines a single self-collected success per comparable task--object--scene case with RGB-only perception, training-free deployment, confidence gating, and guidance-based execution.
By transferring endpoint geometry rather than scene-specific actions, it enables sparse memory to remain useful under memory--scene mismatch.

\section{Implementation Details for Target World-Frame Position}
\label{app:relative_position}

This section describes how the target mask $M_{tar}$ provides
the position $\hat p_{tar}$ and orientation reference
$\hat q_{tar}$ for relational matching and transfer used in Eq.~\ref{eq:world_to_gripper}.
The same procedure is applied to memory observations and
online queries.

\paragraph{Standard calibration.}
For the real robot, camera intrinsics $K$ and distortion parameters
are estimated using a ChArUco board with planar camera calibration
\citep{zhang2000flexible}, while camera-to-robot alignment is obtained
through standard hand--eye calibration \citep{horaud1995hand}.
To resolve the depth ambiguity of a single RGB observation,
we use the tabletop as a geometric reference.
Observing the calibration board placed flat on the working
surface determines the plane $n^\top p+b=0$,where $p\in\mathbb{R}^3$ is any point on the plane expressed
in the robot-base frame, $n\in\mathbb{R}^3$ is its unit normal,
and $b$ determines its signed offset from the origin.
Intersecting a target's viewing ray with this plane then
provides its estimated world-frame position.

\paragraph{Single-view position.}
To express memory and online observations in a common geometric frame, we estimate object position using calibrated RGB geometry.
Given the distortion-corrected mask centroid $(u_x,u_y)$, camera intrinsics $K$, and world-to-camera extrinsics $[R_c\mid t_c]$, the viewing ray has origin $C=-R_c^\top t_c$ and direction $d=R_c^\top K^{-1}[u_x,u_y,1]^\top$ \citep{hartley2003multiple}.
Intersecting this ray with the calibrated table plane $n^\top p+b=0$ gives
\begin{equation}
\hat p_{tar}
=
C-\frac{n^\top C+b}{n^\top d}\,d.
\label{eq:target_world_position}
\end{equation}
This estimate target object position $p'$ in Eq.~\ref{eq:world_to_gripper}.
Applying the same projection to memory and online observations provides a consistent reference, although elevated objects and viewpoint changes can introduce projection bias.

\paragraph{Normalized relational distance.}
\label{app:relation_distance}
Retrieval should reflect geometric compatibility without being dominated by coordinate units or rotation parameterization.
We therefore define the residual used in Eq.~\ref{eq:retrieval} as
\begin{equation}
\|r_a\ominus r_b\|_2
=
\sqrt{
\frac{\|p_a^{\mathrm{rel}}-p_b^{\mathrm{rel}}\|_2^2}{s_p^2}
+
\frac{\theta_{ab}^2}{s_R^2}
},
\label{eq:relation_distance}
\end{equation}
where $\theta_{ab}=\|\operatorname{Log}((Q_a^{\mathrm{rel}})^\top Q_b^{\mathrm{rel}})\|_2$ measures angular separation and $Q^{\mathrm{rel}}=R(q_t)^\top R(q')$.
For undirected planar axes, we instead use the smallest relative-angle difference modulo $\pi$.
The scales $s_p$ and $s_R$ are calibrated as the root-mean-square positional and angular differences between valid states from distinct successful trajectories of the same task and stage, then fixed during evaluation.
This balances the two terms without a manually chosen angular weight; unavailable or zero-variance terms are omitted.

\paragraph{Object orientation.}
To transfer a grasp relation across changes in object alignment,
we estimate the target's planar orientation from the principal
axis of its mask.
Let $\bar u_{tar}$ be the mask centroid and $v_1$ the principal
eigenvector of its pixel covariance.
We project two points $\bar u_{tar}\pm\alpha v_1$, with fixed
pixel span $\alpha$, onto the reference plane using the same
distortion correction and camera geometry as above.
The resulting world-frame points $\hat p^\pm$ define
\begin{equation}
\hat\theta_{tar}
=
\operatorname{atan2}
\bigl(\hat p^+_y-\hat p^-_y,\,
      \hat p^+_x-\hat p^-_x\bigr),
\qquad
\hat q_{tar}=[0,0,\hat\theta_{tar}]^\top.
\end{equation}
Here, $\hat q_{tar}$ supplies the yaw-only object-orientation
reference $q'$ in Eq.~\ref{eq:world_to_gripper}.
The angular difference between the current and memory axes
rotates the retrieved gripper orientation, preserving its
alignment relative to the object rather than its absolute
world-frame direction.
Since the axis is undirected, angular comparisons treat
orientations separated by $\pi$ as equivalent.

\paragraph{Two-view refinement.}
To reduce the localization bias caused by projecting elevated
objects onto the tabletop, we use a second view when both the
fixed and wrist cameras observe the target.
Their mask centroids define world-frame rays
$C_m+\rho_m d_m$, $m\in\{1,2\}$, where $C_m$ is the camera
center, $d_m$ the viewing direction, and $\rho_m$ the position
parameter along the ray.
We estimate the target position from the midpoint of the
closest points on these rays following \cite{hartley2003multiple}:
\begin{equation}
\begin{aligned}
(\rho_1^\star,\rho_2^\star)
&=
\arg\min_{\rho_1,\rho_2}
\left\|
(C_1+\rho_1d_1)-(C_2+\rho_2d_2)
\right\|_2^2,\\
\hat p_{tar}
&=
\frac{
(C_1+\rho_1^\star d_1)
+
(C_2+\rho_2^\star d_2)
}{2},
\end{aligned}
\label{eq:target_two_view}
\end{equation}
where $\rho_1^\star$ and $\rho_2^\star$ locate the closest
points.
The wrist-camera pose is updated from the gripper pose
$(p_t,q_t)$ and the fixed hand--eye transform, requiring
no recalibration as the arm moves.
We retain the single-view estimate when visibility or ray
geometry is unreliable, using the additional view to refine
localization without making it a prerequisite.
This refinement remains approximate because mask centroids
need not correspond to the same physical point across views.

\section{Alternative Sampling-based Reranking for Future-Guided Generation}
\label{app:rerank}

For policies without discrete action logits, FG can guide execution by selecting among complete action proposals.
This alternative preserves the frozen policy's generated actions while using the retrieved target to identify a promising continuation.
Guidance is applied after generation, introducing additional sampling and evaluation cost.

\textbf{Geometric selection.}
The frozen policy generates $N$ candidate chunks $\{a^n_{1:H}\}_{n=1}^{N}$ using independent noise initializations for diffusion or flow-matching heads, with the original policy output retained as one candidate.
To assess their progress toward $r^\star$, we predict the gripper pose along each candidate's complete action prefixes according to its control interface.
Cartesian actions update the pose through commanded displacements, while joint actions determine configurations evaluated through forward kinematics.
At each step $h$, the predicted pose and the fixed current object estimate yield $\hat r_h^n$ through Eq.~\ref{eq:world_to_gripper}.
We jointly select the candidate and execution length that minimize the error:
\begin{equation}
(n^\star,h^\star)
=
\arg\min_{\substack{1\leq n\leq N\\1\leq h\leq H_s}}
\left\|\hat r_h^n\ominus r^\star\right\|_2^2,
\label{eq:rerank_select}
\end{equation}
where $H_s=H$ during localization and $H_s=H'$ during interaction, and $\ominus$ is the normalized geometric residual used in retrieval.

When CG enables geometric selection, RAAG executes the prefix $a^{n^\star}_{1:h^\star}$ unchanged, preserving the coordinated actions within a single proposal.
It then obtains a new observation and replans.
We use $N=3$, trading additional computation for guidance without modifying the continuous-action head.

\textbf{Policy confidence for flow-based heads.}
To determine whether geometric selection is warranted, CG estimates policy risk from the frozen multimodal representations that condition the action expert.
We obtain $z^{cur}$ by mean-pooling the final-layer representations over valid observation and instruction tokens, excluding padding and action-suffix tokens.
Because this embedding is computed before sampling, the risk estimate is independent of the noise used to generate individual proposals.

Combining this estimate with retrieval confidence as in Sec.~\ref{sec:gate} yields $\beta_t$.
We apply geometric selection when $\beta_t/\beta_{\max}\geq0.2$.
Otherwise, we execute the original policy output with its native execution horizon, without reranking or target-based prefix truncation.
Thus, the same confidence estimates that modulate logit-based guidance determine whether retrieved geometry influences proposal selection.

\section{Additional Implementation Details}
\label{app:implementation}

\subsection{Benchmarks}
\label{app:benchmarks}
RAAG is evaluated as a test-time plug-in with all VLA weights frozen.
Each backbone follows its released protocol, while the memory bank and CG thresholds are built only from successful in-distribution rollouts and fixed before evaluation.

\textbf{LIBERO \citep{liu2023libero}.}
LIBERO is one of the most widely used benchmarks for VLA evaluation.
Following the standard protocol, we evaluate $40$ tasks across four $10$-task suites: Spatial, Object, Goal, and Long, using both third-person and wrist-camera observations.
The first three test spatial, object, and goal generalization, while Long emphasizes longer-horizon manipulation.
For each suite, RAAG constructs its memory bank exclusively from successful rollouts.
For LIBERO-Long, we manually segment successful trajectories into ordered subtasks and store their memories separately.
RAAG follows this predefined order and restricts retrieval to the current subtask, within which geometric matching determines the object to manipulate and the placement target.
Repeated retrieval remains within the same subtask until execution advances to the next.

\textbf{LIBERO-Plus \citep{fei2026libero}.}
LIBERO-Plus extends LIBERO into a large-scale robustness benchmark with $10{,}030$ perturbed task instances across seven factors: camera, robot state, language, lighting, background, sensor noise, and layout.
We evaluate all factors using the \textbf{same} LIBERO memory bank, without adding any perturbed rollouts.
For camera-view perturbations, we obtain the updated camera calibration from the simulator for geometric estimation.
This directly tests whether fixed in-distribution experience transfers under broad visual, geometric, and semantic shifts.

\textbf{SimplerEnv Bridge \citep{li2024simplerenv}.}
SimplerEnv Bridge tests portability to a new simulator and robot.
We evaluate four visual-matching tasks using a single third-person camera view, with $3$ runs and $24$ episodes each, totaling $288$ episodes per backbone.
Memory contains only successful executions from the official configurations and remains fixed during evaluation.

\textbf{ManiSkill3 \citep{tao2025maniskill3}.}
We use \textsc{PutOnPlate25}, a $25$-task pick-and-place benchmark with diverse objects and receptacles, using a single third-person camera view.
The adopted protocol contains one in-distribution setting and $13$ OOD conditions covering appearance changes, unseen objects, receptacles and instructions, and altered object or robot states.
One bank per model, collected from that checkpoint's own successful ID rollouts and fixed across all OOD conditions.

\textbf{Our FailureBench.}
Recovering from genuine VLA failures is substantially harder than acting from clean initial states.
The policy must continue from a state produced by an unsuccessful rollout, while no failed trajectories or recovery demonstrations are available. 
We therefore introduce \textsc{FailureBench}, in which each policy starts from the exact simulator state where a frozen VLA previously failed. 
Moreover, the memory bank contains only successful trajectories, directly testing whether successful experience can transfer to unseen failure states rather than simply replaying near-matched recovery behavior.

\begin{figure}[ht]
\centering
\includegraphics[width=0.7\linewidth]{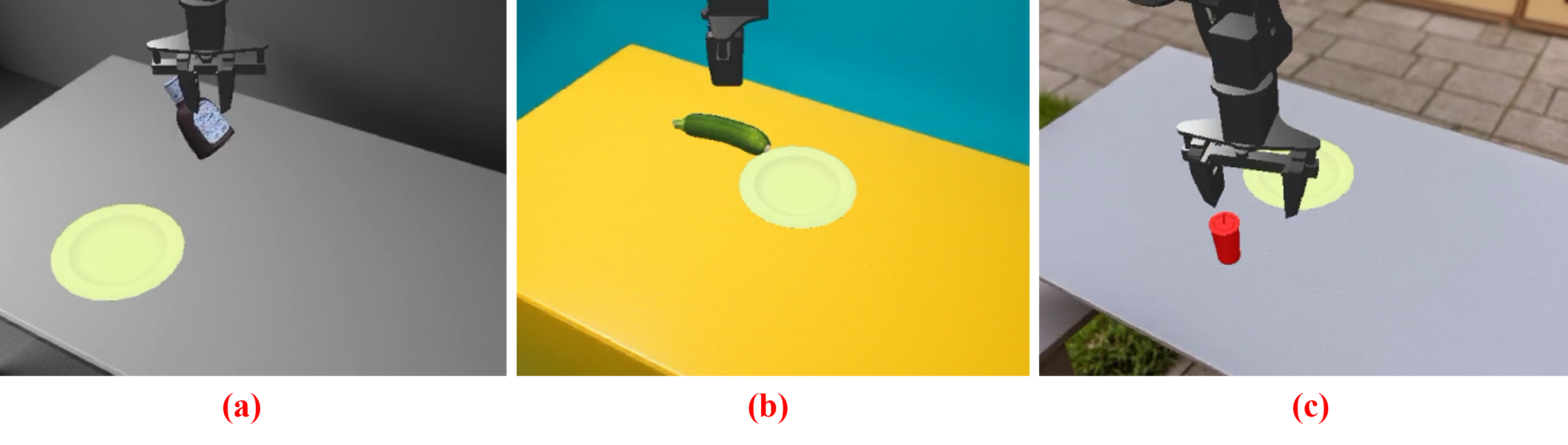}
\caption{\textbf{Examples of our FailureBench.}
(a) Appearance change. (b) Adjacent object. (c) Unusual position.}
\label{fig:failure_bench_example}
\end{figure}

Built from OpenVLA-OFT \citep{kim2025openvlaoft} failures across the $14$ ManiSkill3 \textsc{PutOnPlate25} suites, \textsc{FailureBench} is constructed by running OFT-base with an $80$-step budget and retaining the terminal states of unsuccessful episodes. We exclude physically unrecoverable cases, such as objects knocked off the table, resulting in $856$ recoverable states. As illustrated in Fig.~\ref{fig:failure_bench_example}, simulator ground truth groups these states into three failure types: \emph{appearance change}, where the target is toppled or flipped; \emph{adjacent object}, where it touches or closely neighbors another object; and \emph{unusual position}, where it lies far from its typical training location or presents a difficult grasp pose. We evaluate $856$ states, comprising $207$, $328$, and $321$ cases of these three types, respectively.

\subsection{Run-to-Run Variation}
We repeat three times to estimate evaluation noise. 
The observed mean absolute deviation is $0.9\%$ on FailureBench, $0.4\%$ on the ManiSkill3, $0.4\%$ on LIBERO, and $0.3\%$ on LIBERO-Plus.

\subsection{Memory bank}
\label{app:memory_bank}
To make the memory budget explicit, Tab.~\ref{tab:memory_bank_size} reports the number of trajectories used for each benchmark, which is substantially smaller than in prior retrieval-based methods.
Each trajectory records one successful rollout, including its task instruction, RGB observations, and executed actions at each step.
RAAG further precomputes the target mask and gripper--object relation at each frame and stores the action-conditioning embeddings used by CG to improve the test-time efficiency.

\begin{table*}[ht]
\centering
\caption{\textbf{Number of memory trajectories used for each benchmark.} LIBERO-Plus reuses the memory bank of LIBERO.}
\label{tab:memory_bank_size}
\resizebox{\textwidth}{!}{
\begin{tabular}{lccccccc}
\toprule
\textbf{Benchmark}
& \textbf{ManiSkill3}
& \textbf{FailureBench}
& \textbf{LIBERO-Spatial}
& \textbf{LIBERO-Object}
& \textbf{LIBERO-Goal}
& \textbf{LIBERO-Long}
& \textbf{SimplerEnv Bridge} \\
\midrule
\textbf{Memory size}
& 256
& 256
& 170
& 152
& 40
& 236
& 37  \\
\bottomrule
\end{tabular}
}
\end{table*}

\subsection{Policies}
\label{app:policies}
We evaluate RAAG with three frozen policy backbones:
OpenVLA-OFT \citep{kim2025openvlaoft}, $\pi_0$ \citep{black2024pi_0}, and $\pi_{0.5}$ \citep{intelligence2025pi_}.
The flow-based $\pi$ policies use the sampling-based FG variant described in Appx.~\ref{app:rerank}.

For official OpenVLA-OFT, we evaluate both a discrete action head and a continuous L1-regression head. 
The L1 head achieves higher performance at comparable inference cost \citep{kim2025openvlaoft}, but does not expose action probabilities for direct logit-bias guidance.
Applying RAAG to this head would therefore require the more computationally expensive sampling-based variant.
We instead apply RAAG to the discrete head, which supports efficient probability-level guidance, and report unguided results for both heads to separate the benefit of guidance from the action representation.

On ManiSkill3, we additionally evaluate the GRPO \citep{shao2024deepseekmath}, and PPO \citep{schulman2017proximal} OpenVLA-OFT checkpoints released by RLinf \citep{zang2025rlinf}, spanning different levels of policy competence.
We construct separate memory banks from each model's successful rollouts.

\section{Real-Robot Evaluation}
\label{app:real_robot}

We evaluate whether RAAG can improve a frozen policy's real-world
pick-and-place performance under varied object placements and perturbations. 
To isolate the effect of retrieval guidance, we use the same demonstration 
collection and policy checkpoint for the comparisons with and without RAAG.

\textbf{Robot setup.}
We use the open-source LeRobot SO-ARM101, with STS3250 servos replacing the
shoulder-lift and elbow-flex servos to increase payload capacity, while the remaining
joints retain standard STS3215 servos. A fixed overhead RGB camera provides
workspace context, while a wrist-mounted RGB camera provides a close view of
the grasp region. To connect visual observations to geometry,
we calibrate the camera intrinsics and camera-to-robot transforms using a
ChArUco board, together with the table plane and encoder-to-URDF joint offsets.
This establishes the common coordinate system needed to compare remembered
gripper--object relations with poses predicted by forward kinematics.

\begin{figure*}[t]
\centering
\includegraphics[width=\textwidth]{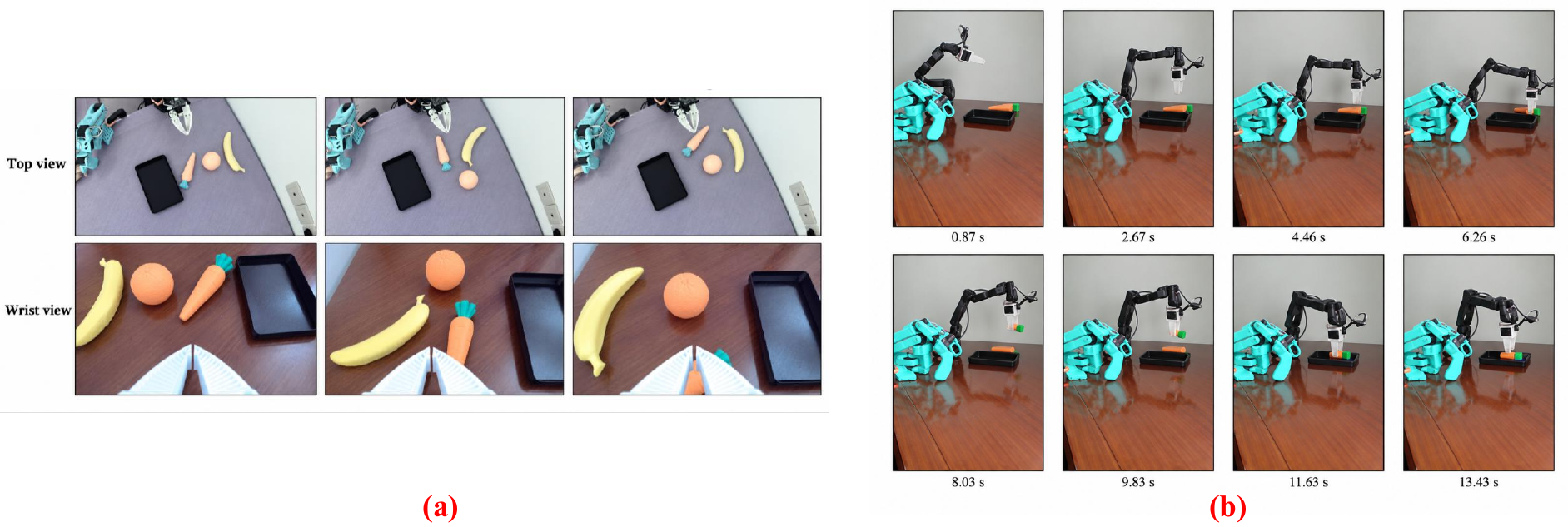}
\caption{\RAAGchange{Real-robot demonstrations.}
\RAAGchange{\textbf{(a)} Three training examples}
shown from the overhead and wrist cameras. In each
example, one object is the instructed grasp target and the remaining objects
serve as distractors.
\RAAGchange{\textbf{(b)} An example execution with progression timestamps.}}
\label{fig:real_robot_setup_examples}
\end{figure*}

\textbf{Objects and demonstrations.}
We collect 40 successful demonstrations for each of three open-source
3D-printed objects---a carrot, a banana, and an orange---yielding 120
trajectories recorded at 10\,Hz through leader--follower teleoperation.
The spherical orange is particularly sensitive to grasp alignment,
as it can easily slip out as the gripper closes, demanding precise
localization and positioning.
Initial placements vary, and each scene contains an instructed target
alongside two distractors as shown in Fig.~\ref{fig:real_robot_setup_examples}(a). 
The instruction, \emph{Put the [object] on the black plate}, 
therefore requires selecting a specific object rather than simply
grasping an available one. 
We use those demonstrations to both fine-tune the base policy and construct
RAAG's memory bank, giving RAAG no additional demonstration data.
This allows us to test whether retrieving information from demonstrations
already used for training can further improve the policy's robustness
across different test conditions when demonstrations are limited.

\textbf{Policy training.}
We initialize from OpenVLA-7B~\citep{kim2024openvla} and adapt the official
OpenVLA-OFT~\citep{kim2025openvlaoft} training implementation to the SO-ARM101.
The policy takes the instruction, two $256\times256$ RGB observations,
and the robot's proprioceptive state. Feature-wise Linear Modulation
(FiLM; \citealp{perez2018film}) conditions visual features on the instruction
to support target selection in multi-object scenes.

We use the parallel discrete-action variant, which exposes action
probabilities for RAAG's decoding guidance. Each query predicts a 10-step
chunk of Cartesian end-effector actions and gripper commands, with 256 bins
per action dimension. Demonstration actions are converted to end-effector
space using forward kinematics. At deployment, RAAG guides the Cartesian
action predictions, which are converted into joint commands through
LeRobot's URDF-based inverse-kinematics interface.
We fine-tune with action-token cross-entropy, rank-32 LoRA, and AdamW
for 20 epochs, using a learning rate of $5\times10^{-4}$ and an effective
batch size of 64 following the standard practice \citep{kim2025openvlaoft}. 
Both evaluation methods use the same frozen checkpoint
and inverse-kinematics controller.

\textbf{RAAG configuration.}
RAAG uses SAM~3~\citep{carion2026sam3} to obtain object and plate
masks from the overhead observation.
Following the detect-then-track strategy of Grounded SAM \citep{ren2024grounded},
we perform open-vocabulary segmentation on the first frame and
propagate the masks through subsequent frames, maintaining temporal
correspondence without repeating open-vocabulary detection.

To use these observations for geometric retrieval, we back-project
the masks onto the calibrated table plane to estimate object and
plate coordinates in the robot base frame, and obtain the gripper
pose through forward kinematics.
Expressing demonstrations and test observations in this common frame
allows RAAG to compare their gripper--object relations and retrieve
a compatible successful configuration.
The retrieved relation is then instantiated at the object's current
location to guide the policy's action predictions.
In this way, a demonstration collected at one placement can provide
useful guidance at another, without replaying its original joint
trajectory.

\textbf{Evaluation protocol.}
We evaluate whether RAAG improves both task execution under familiar
conditions and generalization to an unseen placement target.
All demonstrations use a black plate, whereas evaluation includes
both black and white plates. The white-plate condition changes the
target's appearance while preserving the pick-and-place objective,
allowing us to examine whether guidance based on geometric relations
remains useful beyond the demonstrated target appearance.
The instruction specifies the plate color in each condition.

For each object and plate, we evaluate 20 distinct initial positions with
and without RAAG. We alternate the two methods and restore the initial scene
as closely as possible between paired attempts to reduce the
influence of scene variation and temporal drift.
Both methods share the same observation processing and robot
execution settings. Success requires the robot to grasp the
instructed object and release it onto the specified plate, where
it remains supported without human assistance.

\textbf{Results.}
Table~\ref{tab:real_robot_results} shows that RAAG improves success
under both plate conditions, with the largest gain on the unseen
white plate: from 10.0\% to 56.7\%.
The substantial degradation of the direct policy under this
unseen target condition suggests sensitivity to visual features specific
to the training setting. By providing a geometric prior retrieved
from demonstrations, RAAG reduces this sensitivity and supports
more robust transfer to the unseen plate.

RAAG also shortens completion time on successful trials with only
3.4\,ms of additional decision latency. This is consistent with
more accurate localization reducing corrective motions and repeated
grasp attempts, although the successful trial subsets differ
between methods.

\begin{table*}[t]
\centering
\caption{Real-robot pick-and-place results on seen black plates and
unseen white plates. Both methods use the same discrete OpenVLA-OFT
checkpoint trained exclusively on black-plate demonstrations.
We report success rate (\%),
with overall results pooled across the three objects.
Completion time is measured on successful trials, while decision
latency is measured across all policy queries.}
\label{tab:real_robot_results}
\setlength{\tabcolsep}{4pt}
\resizebox{0.95\textwidth}{!}{%
\begin{tabular}{llcccccc}
\toprule
Target plate & Method & Carrot & Banana & Orange & Overall
& \shortstack{Completion\\time (s) $\downarrow$}
& \shortstack{Decision\\latency (ms) $\downarrow$} \\
\midrule
\multirow{2}{*}{\shortstack[l]{Black\\(seen)}}
& OpenVLA-OFT        & 45 & 40 & 15 & 33.3 & 19.3 & 174.2 \\
& OpenVLA-OFT + RAAG & 80 & 70 & 35 & 61.7 & 15.8 & 177.6 \\
\midrule
\multirow{2}{*}{\shortstack[l]{White\\(unseen)}}
& OpenVLA-OFT        & 20 & 10 &  0 & 10.0 & 22.5 & 174.2 \\
& OpenVLA-OFT + RAAG & 75 & 70 & 25 & 56.7 & 20.6 & 177.6 \\
\bottomrule
\end{tabular}}
\end{table*}

\begin{RAAGchangeblock}
\subsection{Sim-to-Real Memory Transfer}
\label{app:sim_to_real_memory}

We conduct an additional carrot-placement experiment in a new real-world environment to test whether simulation memories can guide a real robot.
OpenVLA-OFT is fine-tuned on $20$ real-robot carrot-placement demonstrations and then frozen.
All three conditions use this checkpoint and are evaluated for $20$ trials each in the same environment as policy training.
RAAG uses either carrot-placement memories collected in ManiSkill3 or real-world memories; the simulation-memory condition adds no real-world memories.
The retrieved gripper--object relation specifies a grasp goal relative to the current carrot, as illustrated in Fig.~\ref{fig:sim_to_real_memory}.

\begin{figure}[t]
\RAAGchangecolor
\centering
\begin{minipage}[c]{0.38\textwidth}
\centering
\footnotesize
\setlength{\tabcolsep}{3pt}
\renewcommand{\arraystretch}{1.15}
\begin{tabular}{@{}lr@{}}
\toprule
Method & Success (\%) \\
\midrule
VLA & 55 (11/20) \\
\ourhl VLA + RAAG (sim mem.) & \textbf{70 (14/20)} \\
VLA + RAAG (real mem.) & 75 (15/20) \\
\bottomrule
\end{tabular}
\captionof{table}{\textbf{Sim-to-real memory transfer.} All conditions share the frozen checkpoint and use $20$ real-robot trials each.}
\label{tab:sim_to_real_memory}
\end{minipage}\hfill
\begin{minipage}[c]{0.60\textwidth}
\centering
\includegraphics[width=\linewidth]{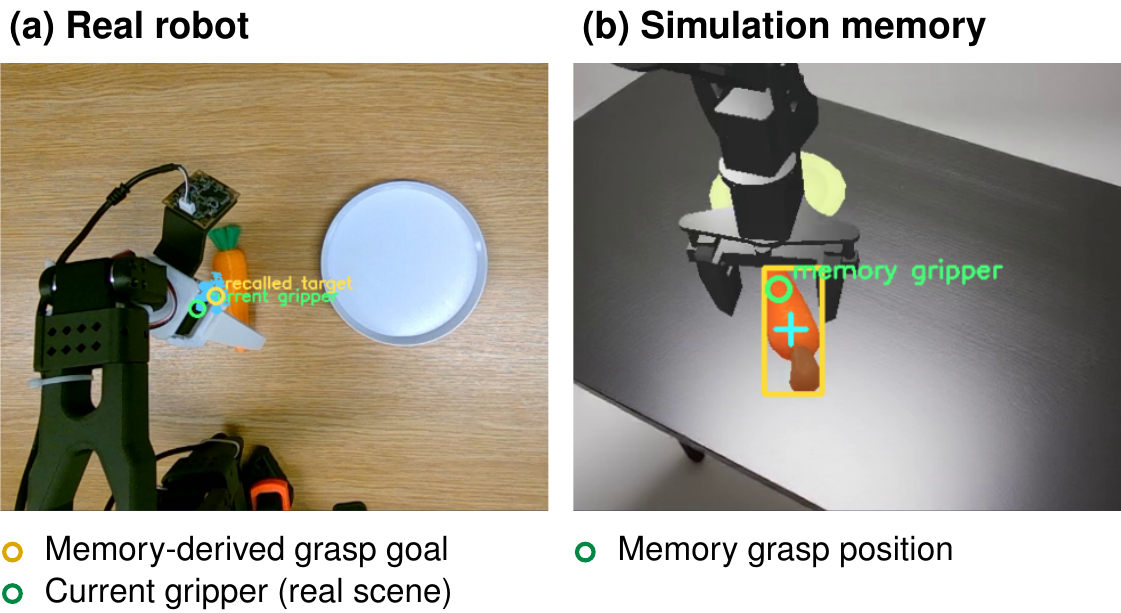}
\captionof{figure}{\textbf{Using simulation memory on a real robot.} Left: the yellow marker is the memory-derived goal; green is the current gripper. Right: the retrieved ManiSkill3 endpoint.}
\label{fig:sim_to_real_memory}
\end{minipage}
\end{figure}

Simulation memory raises success from $55\%$ to $70\%$, approaching the $75\%$ obtained with real-world memory (Tab.~\ref{tab:sim_to_real_memory}).
These preliminary results show that useful gripper--object geometry can transfer from simulation to a real robot without additional policy training or new real-world memories.
This supports a practical, reproducible way for the community to improve robot generalization through accumulated experience, reducing the need for new demonstrations and policy retraining for every scene.
\end{RAAGchangeblock}

\section{Inference Cost}
\label{app:efficiency}

RAAG's efficiency comes from overlapping memory access with policy inference rather than adding a sequential future-generation stage.
As shown in Table~\ref{tab:efficiency}, the retrieval pipeline takes $52.6$\,ms, primarily spent on object segmentation with the SAM~3~\citep{carion2026sam3} perception foundation model, and runs concurrently with the $147.3$\,ms OpenVLA-OFT forward pass on our dual-GPU system.
Only guidance and validation remain sequential, adding $4.2$\,ms in total and yielding a decision latency of $151.5$\,ms---a $2.9\%$ increase over the base policy.
RAAG thus supplies an experience-based geometric prior without training a world model or explicitly generating future observations.

In real-world deployment, SAM~3 perception runs on the retrieval branch. 
Its cost can likewise be hidden when this branch completes before policy inference.
This small critical-path overhead therefore depends on concurrent execution: on a single GPU, resource contention or serialized execution can expose additional perception and retrieval costs.

\begin{table}[!t]
\centering
\caption{\textbf{Per-decision latency breakdown} in milliseconds.
$\parallel$ denotes stages executed in parallel with the VLA forward pass, so Total follows the critical path.
WM data denotes the samples used to train a world model or failure supervisor. 
"Generation" records the additional latency incurred outside the VLA generation.
Parentheses denote serialized/single-GPU latency.}
\label{tab:efficiency}
\setlength{\tabcolsep}{5pt}
\renewcommand{\arraystretch}{1.05}
\resizebox{0.8\textwidth}{!}{%
\begin{tabular}{@{}l c ccccc@{}}
\toprule
\textbf{Method} & \textbf{WM data} & \textbf{VLA} & \textbf{Retrieve} & \textbf{Generation} & \textbf{Gating} & \textbf{Total} \\
\midrule
\basehl OpenVLA-OFT (frozen) \citep{kim2025openvlaoft} & 0 & 147.3 & --- & --- & --- & 147.3 \\
\midrule
\wamhl VLA-JEPA \citep{sun2026vla} & 296K & 74 & --- & --- & --- & 74 \\
\wamhl CoT-VLA \citep{zhao2025cot} & 311K & 108 & --- & 5845 & --- & 5953 \\
\wamhl DreamVLA \citep{zhang2026dreamvla} & 76K & 88 & --- & 3 & --- & 91 \\
\wamhl FPC-VLA \citep{yang2026fpc} & 110K & 180 & --- & 1590 & --- & 1770 \\
\midrule
\retrhl $\mu$VLA \citep{cherepanov2026mu} & 0 & 143.4 & --- & 6.0 & --- & 150.4 \\
\retrhl OptimusVLA \citep{li2026global} & 0 & 179.2 & --- & 3.5 & --- & 182.7 \\
\midrule
\ourhl \method{} (ours) & 0 & 147.3 & 52.6$^{\parallel}$ & 1.3 & 2.9 & 151.5$^{\parallel}$ (204.1) \\
\bottomrule
\end{tabular}}
\end{table}

\section{Retrieval Stability and Sensitivity}
\label{app:sensitivity}

\begin{table}[!t]
\centering
\small
\caption{\textbf{Sensitivity to memory budget and retrieval parameters} on ManiSkill3.}
\label{tab:appx:sens}

\begin{minipage}[t]{0.46\linewidth}
\centering
\begin{tabular}{@{}lcc@{}}
\toprule
& FailureBench & ManiSkill3 \\
\midrule

\multicolumn{3}{@{}l}{\emph{(A) Memory budget}} \\
None
& 13.1
& 24.0 \\
1 case
& 3.8
& 9.5 \\
16 cases
& 20.1
& 47.8 \\
32 cases
& 27.5
& 56.0 \\
64 cases
& 31.4
& 62.9 \\
128 cases
& 34.3
& 66.3 \\
\textbf{256 cases} 
& \textbf{39.1}
& \textbf{69.6} \\

\bottomrule
\end{tabular}
\end{minipage}
\hfill
\begin{minipage}[t]{0.46\linewidth}
\centering
\begin{tabular}{@{}lcc@{}}
\toprule
& FailureBench & ManiSkill3 \\
\midrule

\multicolumn{3}{@{}l}{\emph{(B) Retrieval parameters}} \\
$\delta=1$ 
& 31.4
& 69.2 \\
\textbf{$\delta=$ 2  }
& \textbf{39.1}
& \textbf{69.6} \\
$\delta=4$ 
& 23.9
& 56.5 \\
$w_s=0$
& 33.8
& 64.0 \\
\textbf{$w_s=$ 0.4}
& \textbf{39.1}
& \textbf{69.6} \\
$w_s=0.8$
& 33.9
& 69.2 \\
\bottomrule
\end{tabular}
\end{minipage}
\end{table}

In addition to Sec.~\ref{sec:exp}, we further examine whether RAAG retrieves stable memories using different memory banks and hyperparameters in Tab.~\ref{tab:appx:sens}.

\textbf{Memory budget.}
RAAG benefits from a small but sufficiently diverse memory bank.
With only one case, limited interaction coverage can produce mismatched targets, while CG lacks the cross-trajectory samples needed for reliable calibration.
Unreliable guidance can therefore override useful policy predictions, consistent with success dropping from $13.1/24.0$ to $3.8/9.5$.
Expanding the bank to just $16$ cases raises success to $20.1/47.8$, capturing $27\%/52\%$ of the full-bank improvement.
Larger banks generally provide further gains, reaching $39.1/69.6$ with $256$ cases.
These results suggest that RAAG needs sufficient coverage for compatible retrieval and confidence calibration, but can transfer successful relations without densely storing near-matched trajectories.

\textbf{Endpoint precision and object shape.}
A tight endpoint is essential.
Relaxing $\delta$ from $2$ to $4$ reduces success from $39.1/69.6$ to $23.9/56.5$, while an overly strict $1$ threshold also hurts FailureBench.
Removing shape matching reduces performance by $5.3$ and $5.6$ points, whereas increasing its weight to $0.8$ provides no further benefit.
These results show that geometry is the primary retrieval signal, while object shape serves as a complementary cue for selecting an interaction-compatible relation.
The deployed values were fixed using in-distribution memory and reused across all benchmarks without per-benchmark tuning.


\section{Mechanism Analysis on Confidence Gating}
\label{sec:embood}

\begin{table*}[!t]
\centering

\begin{minipage}[t]{0.38\textwidth}
\centering
\caption{\textbf{Failure-prediction AUROC of the two CG signals.}}
\label{tab:embood}

\resizebox{\linewidth}{!}{%
\begin{tabular}{llcc}
\toprule
Setting & Policy & $D_{\mathrm{emb}}$ & $D_{\mathrm{rel}}$ \\
\midrule
\multirow{3}{*}{ManiSkill3}
 & Base & \textbf{0.71} & 0.52 \\
 & GRPO & \textbf{0.89} & 0.46 \\
 & PPO  & \textbf{0.88} & 0.83 \\
\midrule
\multirow{3}{*}{FailureBench}
 & Base & \textbf{0.76} & 0.48 \\
 & GRPO & \textbf{0.80} & 0.47 \\
 & PPO  & \textbf{0.82} & 0.70 \\
\bottomrule
\end{tabular}}
\end{minipage}
\hfill
\begin{minipage}[t]{0.36\textwidth}
\centering
\caption{\textbf{Sensitivity to the nearest neighbors for confidence gating.}}
\label{tab:knn-sensitivity}

\resizebox{\linewidth}{!}{%
\begin{tabular}{ccccc}
\toprule
$k$ & Base & GRPO & PPO & Avg. \\
\midrule
0  & \textbf{71.0} & 74.3 & 81.9 & 75.7 \\
1  & 67.6 & 76.8 & \textbf{91.4} & 78.6 \\
2  & 69.6 & \textbf{80.5} & 90.0 & \textbf{80.0} \\
3  & 66.7 & 74.1 & 83.3 & 74.7 \\
4  & 63.3 & 70.8 & 76.5 & 70.2 \\
5  & 61.0 & 67.1 & 59.7 & 62.6 \\
\bottomrule
\end{tabular}}
\end{minipage}
\end{table*}

RAAG should intervene only when the frozen policy is unreliable.
Thus, CG must judge whether an action proposal is likely to fail without training a separate detector.
We evaluate two signals by replacing $z^{cur}$ and $\mathcal{Z}_{succ}$ in Eq.~\ref{eq:vla_confidence}.
The embedding signal uses the frozen VLA representation that jointly encodes the observation, robot state, and instruction and directly conditions action generation.
A large distance from successful embeddings therefore indicates an unfamiliar and potentially unreliable proposal.
The relational signal instead compares the gripper-target relation induced by the proposal with successful interaction relations, directly detecting geometrically abnormal actions.

Tab.~\ref{tab:embood} shows that embedding distance is the stronger general risk signal, reaching AUROC $0.89/0.88$ on fresh OOD rollouts and $0.80/0.82$ on genuine failure states for GRPO/PPO.
Relational distance is weaker overall because it captures only interaction geometry, which varies with the task and initial robot state, whereas embedding distance jointly accounts for these factors.
We therefore use embedding distance to estimate general policy risk and retrieval distance to detect geometrically unreliable proposals.

In Tab.~\ref{tab:knn-sensitivity}, we further vary the number of nearest successful embeddings $K$ (used to compute $D_{\mathrm{emb}}$ in Eq.~\ref{eq:vla_confidence}) for policies with different competence, where $K{=}0$ disables CG. 
For the weak Base policy, memory guidance is often beneficial, so gating can unnecessarily suppress reliable retrieval. 
In contrast, stronger GRPO and PPO policies benefit from CG by avoiding unnecessary memory intervention, while a large $K$ introduces less relevant success states and weakens the local risk estimate. 
Overall, a small neighborhood ($K{=}2$) provides the best balance across policy strengths.


\section{Cross-Environment Evaluation on SimplerEnv}
\label{app:simpler}

SimplerEnv evaluates whether RAAG transfers beyond its original simulator.
Because OpenVLA-7B achieves limited success in this setting, we construct its retrieval memory from successful VLA-JEPA rollouts.
As shown in Tab.~\ref{tab:simpler-main}, RAAG improves all three backbones, raising OpenVLA-7B from $1.0\%$ to $24.3\%$, SpatialVLA from $42.7\%$ to $50.0\%$, and VLA-JEPA from $57.3\%$ to $66.0\%$.
This result directly supports our central claim that the reusable part of successful experience is the interaction relation rather than the policy-specific action sequence.
RAAG also improves OpenVLA substantially more than V-GPS, another zero-shot plug-in method reported under the same protocol, reaching $24.3\%$ versus $6.5\%$.
These results show that relational guidance transfers across policy architectures and action-generation mechanisms without retraining.

\begin{table}[!htbp]
\centering
\caption{\textbf{Cross-policy transfer on SimplerEnv Bridge.}
Success rate (\%) averaged over three runs, each with 24 episodes per task, across four visual-matching tasks.}
\label{tab:simpler-main}
\setlength{\tabcolsep}{6pt}
\renewcommand{\arraystretch}{1.12}
\resizebox{0.75\textwidth}{!}{%
\begin{tabular}{@{}lccccc@{}}
\toprule
\textbf{Model} & Spoon & Carrot & Stack & Eggplant & \textbf{Avg.} \\
\midrule
\multicolumn{6}{@{}l}{\emph{Frozen zero-shot policies}} \\
OpenVLA-7B \citep{kim2024openvla}
& 0.0 & 0.0 & 0.0 & 4.1 & 1.0 \\
OpenVLA-7B $+$ V-GPS \citep{nakamoto2024steering}
& 0.0 & 4.0 & 2.0 & \textbf{20.0} & 6.5 \\
\textbf{OpenVLA-7B $+$ \method{} (ours)}
& \textbf{41.7} & \textbf{37.5} & \textbf{8.3} & 9.7
& \textbf{24.3} \\
\midrule
SpatialVLA \citep{qu2025spatialvla}
& 16.7 & 25.0 & 29.2 & \textbf{100.0} & 42.7 \\
\textbf{SpatialVLA $+$ \method{} (ours)}
& \textbf{33.3} & \textbf{33.3} & \textbf{33.3}
& \textbf{100.0} & \textbf{50.0} \\
\midrule
VLA-JEPA \citep{sun2026vla}
& \textbf{75.0} & 70.8 & 12.5 & 70.8 & 57.3 \\
\textbf{VLA-JEPA $+$ \method{} (ours)}
& 70.8 & \textbf{77.8} & \textbf{33.3} & \textbf{81.9}
& \textbf{66.0} \\
\bottomrule
\end{tabular}}
\end{table}

\begin{figure}[!t]
\centering
\includegraphics[width=\linewidth]{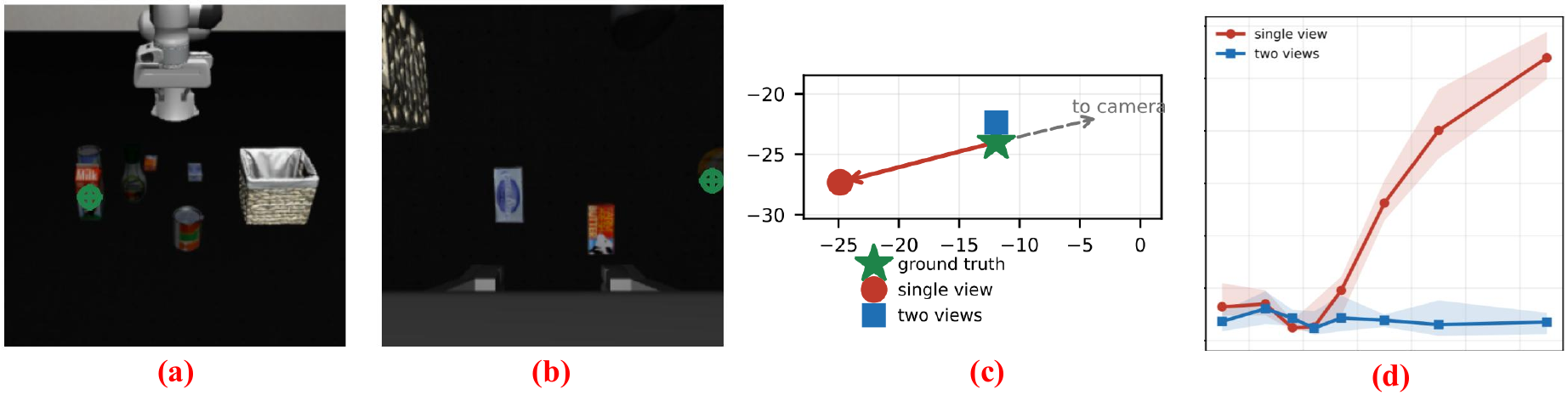}
\caption{\textbf{Single-view depth bias and two-view triangulation.}
(a) A single workspace camera provides only a viewing ray through the target mask centroid.
(b) A wrist camera supplies a second ray, allowing the target position to be triangulated.
(c) An example for top-down localization in world coordinates, where the horizontal and vertical axes show the object's world $x$ and $y$ positions in centimetres. The markers compare the ground-truth position with the single-view and two-view estimates.
(d) Localization error over $5{,}740$ frames. The horizontal axis is the object height bins, and the vertical axis is the resulting 3D localization error in centimetres. The curves show the median error, and the shaded regions indicate the interquartile range.}
\label{fig:view_depth_bias}
\end{figure}

\begin{table}[!t]
\centering
\caption{\textbf{Object localization error under different object heights.}
Median error in centimetres for single-view plane intersection and two-view triangulation on LIBERO frames.}
\label{tab:view_depth_bias}
\setlength{\tabcolsep}{8pt}
\resizebox{0.75\textwidth}{!}{%
\begin{tabular}{@{}lrrrr@{}}
\toprule
Object height above the assumed plane & Frames & Single view & Two views & Reduction \\
\midrule
resting on the table ($<2$\,cm) & 4{,}161 & 1.19 & \textbf{0.30} & 75\% \\
$2$--$5$\,cm                    &   745 & 2.78 & \textbf{0.47} & 83\% \\
$5$--$10$\,cm                   &   285 & 11.12 & \textbf{0.25} & 98\% \\
$>10$\,cm                       &   549 & 22.42 & \textbf{0.18} & 99\% \\
\midrule
all frames                      & 5{,}740 & 2.25 & \textbf{0.29} & 87\% \\
\bottomrule
\end{tabular}}
\end{table}

\section{Failure Analysis}
\label{appx:failure}
RAAG obtains object geometry directly from RGB observations without a trained depth model.
With a single camera, however, a mask centroid defines only a viewing ray.
We therefore intersect this ray with the table plane to estimate the object position.
This is accurate when the object lies on the table, but becomes biased when the visible centroid is above the assumed plane, as shown in Fig.~\ref{fig:view_depth_bias}(c).
A second camera removes this assumption by triangulating the object from two viewing rays.

We compare single-view plane intersection and two-view triangulation on the same $5{,}740$ frames from all four LIBERO suites.
Both methods use the same segmentation centroid and are evaluated against the simulator ground truth, so their difference directly measures the cost of the single-view depth assumption.
As shown in Tab.~\ref{tab:view_depth_bias}, the single-view estimate is reasonable when the object is close to the table, with $1.19$ cm error below $2$ cm height.
Its error then grows rapidly to $11.12$ cm at $5$--$10$ cm and $22.42$ cm above $10$ cm.
In contrast, two-view triangulation remains near $0.30$ cm across all height ranges.
Overall, it reduces the median error from $2.25$ to $0.29$ cm.

Fig.~\ref{fig:view_depth_bias}(d) further shows that the single-view error grows with object height, isolating the main geometric limitation of RAAG.
The transferable relation remains useful, but its retargeting becomes inaccurate when object depth is estimated from an invalid scene assumption.
Multi-view triangulation resolves this failure without learned depth prediction or additional policy training.
However, triangulation requires the object to be visible from both cameras, so single-view plane intersection is retained as a fallback.

%% file: paper_sec/7_perception_execution.tex
\section{SAM 3 Perception Evaluation}
\label{app:sam3}

The main simulation results use simulator instance masks, while the real robot uses off-the-shelf SAM~3 without fine-tuning.
To test predicted masks in simulation, we adapt SAM~3 using only LIBERO in-distribution training-memory data.
Benchmark evaluation trajectories and OOD data are excluded from both fine-tuning and checkpoint selection.

Trajectories are split $7{:}1{:}2$ into training, validation, and test sets.
The vision backbone and text encoder remain frozen; only $24.8$M/$840.4$M parameters (about $3\%$) are updated.
Validation selects the checkpoint at $1{,}500$ steps, using approximately $0.6$ L40 GPU-hours.
The held-out evaluation covers $1{,}419$ object instances from $285$ trajectories.

\begin{table}[ht]
\RAAGchangecolor
\centering
\caption{\textbf{Segmentation evaluation.} Recognition is the percentage of selected target masks with IoU $\geq0.5$ against the correct instance.}
\label{tab:sam3_segmentation}
\small
\begin{tabular}{@{}lrrr@{}}
\toprule
Segmenter & Mean IoU & Recognition (\%) & LIBERO (\%) \\
\midrule
Grounding DINO + SAM~2 & 0.768 & 79.6 & --- \\
Off-the-shelf SAM~3 & 0.844 & 87.0 & 73.4 \\
\ourhl Fine-tuned SAM~3 & \textbf{0.979} & \textbf{99.9} & \textbf{99.8} \\
\bottomrule
\end{tabular}
\end{table}

Fine-tuning raises LIBERO recognition from $73.4$ to $99.8\%$.
The end-to-end results in Tab.~\ref{tab:sam3_tasks} retain RAAG's gains with predicted masks.

%% file: paper_sec/8_controls.tex
\section{Additional Retrieval and Guidance Controls}
\label{app:controls}

\subsection{Fixed-prototype retrieval}
\label{app:prototypes}

We retain perception, successful memory, the frozen checkpoint, stage logic, gating, guidance, and execution frequency; only trajectory selection changes.
The \textbf{relational prototype} is the task-bank medoid: the trajectory minimizing summed normalized differences in approach/placement endpoint positions and approach orientation to the other trajectories.
It uses geometry offline, without shape matching.
The \textbf{object-specific prototype} matches task and target object name, using identity as a shape proxy.
Both fix the trajectory throughout execution, including interaction refinement, while retaining within-trajectory frame matching and endpoint retargeting.

\begin{table}[ht]
\RAAGchangecolor
\centering
\caption{\textbf{Retrieval controls.} Success (\%). Offline geometry compares stored trajectories; online geometry matches the current observation to memory.}
\label{tab:retrieval_controls}
\small
\setlength{\tabcolsep}{4pt}
\begin{tabular}{@{}lcccrr@{}}
\toprule
Memory selection & \makecell{Task/\\stage} & Shape & \makecell{Relation\\matching} & FB & MS3 \\
\midrule
Fixed across verbs & --- & --- & --- & 3.8 & 9.5 \\
Random in task bank & \checkmark & --- & --- & 15.8 & 29.9 \\
Appearance only & \checkmark & \checkmark & --- & 25.5 & 46.2 \\
Relational fixed prototype & \checkmark & --- & Offline & 27.8 & 46.4 \\
Object-specific fixed prototype & \checkmark & Identity & --- & 35.4 & 64.8 \\
\ourhl \textbf{Full GR} & \checkmark & \checkmark & Online & \textbf{39.1} & \textbf{69.6} \\
\bottomrule
\end{tabular}
\end{table}

Task-only selection can choose an unsuitable grasp; shape alone does not ensure a compatible gripper--object configuration.
The relational prototype misses shape-dependent contact requirements.
The object-specific prototype is stronger, but a different object pose can still make its fixed grasp unsuitable, and it requires object-specific memory coverage.
Full GR repeatedly matches shape and current relations, enabling compatible experience to transfer across object identities.

\subsection{Matched geometric guidance}
\label{app:fusion_controls}

Table~\ref{tab:matched_guidance} shares perception, memory, retrieval, endpoints, the frozen OpenVLA-OFT checkpoint, and execution settings.
Let $a$ be the policy translation, $u$ the clipped endpoint-directed step, and $c=c^{ret}$ the shared retrieval confidence.
Rotation and gripper commands remain those of the policy.
\begin{itemize}[itemsep=2pt,topsep=3pt,leftmargin=*]
\item \textbf{Hard switch:} $u$ if $c\geq0.5$, otherwise $a$, following shared-control arbitration \citep{dragan2013policy}.
\item \textbf{Bounded residual:} $a+cu$, capped by the larger of the geometric-step norm and policy-chunk translation norms; an analytic residual-control comparison \citep{johannink2019residual}.
\item \textbf{Minimal projection:} $a+\max\{c\|u\|-a^\top\hat u,0\}\hat u$, where $\hat u=u/\|u\|$; return $a$ when $u=0$.
This tests the minimal-intervention QP form \citep{ames2016control}, without safety guarantees.
\item \textbf{Fixed logit tilt:} the same geometric logit penalty with constant $\beta=30$, analogous to decoding-time reweighting \citep{yang2021fudge}.
\end{itemize}
These controls test operator forms rather than reproducing learned residual policies or full safety-control systems.
FG performs best by combining endpoint compatibility with the policy prior during action selection.

\subsection{Retrieved actions versus endpoint goals}
\label{app:action_payload}

To test the guidance payload, we retain the same retrieval setup and frozen $\pi_{0.5}$, comparing RtS-style recorded-action injection \citep{zhao2026retrieve} with RAAG's endpoint guidance.
RtS-style injection initializes $x_{t_0}=(1-t_0)a^{mem}+t_0\epsilon$ from checkpoint-normalized memory actions and Gaussian noise, then denoises from $t_0$ to zero.
We test $t_0\in\{0.3,0.5,0.7\}$ and report the best, $t_0=0.3$.
This isolates the guidance payload rather than reproducing RtS's complete online-memory system.

RAAG exceeds action injection by $2.7$ points.
A recorded motion may not suit the current arm/object configuration; the endpoint goal lets the motion adapt while preserving the desired interaction relation.
The result supports transferring geometric goals rather than recorded actions in this controlled setting.

\begin{table}[tbp]
\RAAGchangecolor
\centering
\begin{minipage}[t]{0.55\textwidth}
\vspace{0pt}
\centering
\caption{\textbf{Matched geometric guidance.} Success (\%) with shared inputs and frozen OpenVLA-OFT. FB: FailureBench; MS3: ManiSkill3 OOD.}
\label{tab:matched_guidance}
\small
\setlength{\tabcolsep}{3pt}
\begin{tabular}{@{}lrr@{}}
\toprule
Localization operator & FB & MS3 \\
\midrule
Hard switch & 23.7 & 59.8 \\
Bounded residual & 32.4 & 63.4 \\
Minimal projection & 33.8 & 65.1 \\
Fixed tilt, $\beta=30$ & 37.2 & 67.1 \\
\ourhl \textbf{FG with joint CG} & \textbf{39.1} & \textbf{69.6} \\
\bottomrule
\end{tabular}
\end{minipage}\hfill
\begin{minipage}[t]{0.41\textwidth}
\vspace{0pt}
\centering
\caption{\textbf{Action payload on LIBERO-Plus.} Success (\%) with shared retrieval and frozen $\pi_{0.5}$.}
\label{tab:action_payload}
\small
\setlength{\tabcolsep}{3pt}
\begin{tabular}{@{}lr@{}}
\toprule
Guidance & Success \\
\midrule
Frozen policy & 85.7 \\
RtS-style injection & 88.6 \\
\ourhl \textbf{RAAG endpoint} & \textbf{91.3} \\
\bottomrule
\end{tabular}
\end{minipage}
\end{table}

\subsection{Candidate budget and execution horizon}
\label{app:flow_controls}

We evaluate the same $240$ LIBERO-Plus episodes, $80$ each from Cam/Rob/Lang, under official episode limits.
Horizon is the maximum executed prefix before replanning; samples counts candidate action chunks.
Geometric selection chooses an unchanged policy candidate and prefix closest to the endpoint (Eq.~\ref{eq:rerank_select}).

\begin{table}[ht]
\RAAGchangecolor
\centering
\caption{\textbf{Sampling and horizon controls.} Success (\%). 10/2 caps localization at ten actions and interaction at two; geometric prefix selection can stop earlier.}
\label{tab:flow_controls}
\small
\setlength{\tabcolsep}{5pt}
\begin{tabular}{@{}lcccr@{}}
\toprule
Arm & Horizon & Samples & \makecell{Geometric\\selection} & Success \\
\midrule
Frozen $\pi_{0.5}$, native & 10 & 1 & --- & 79.6 \\
Frozen, equal candidate budget & 10 & 3, random & --- & 80.0 \\
Frozen, short horizon & 2 & 1 & --- & 82.5 \\
\ourhl Full RAAG & 10/2 & 3 & \checkmark & \textbf{85.8} \\
RAAG, short horizon & 2 & 3 & \checkmark & 86.7 \\
\bottomrule
\end{tabular}
\end{table}

RAAG reaches $85.8\%$ versus the short-horizon policy's $82.5\%$, while permitting longer localization prefixes.
Random extra samples reach only $80.0\%$.
Further shortening RAAG's horizon adds $0.9$ points but requires more frequent replanning.
The gains therefore extend beyond extra sampling or a shorter configured horizon alone.

%% file: paper_sec/9_calibration.tex
\section{Camera Calibration and Self-Calibration}
\label{app:self_calibration}

RAAG uses calibration, masks, and proprioception to construct gripper--object geometry.
Main simulation results use exact simulator intrinsics/extrinsics, including updated LIBERO-Plus camera parameters.
Image-only VLA baselines have no camera-matrix input; geometric controls receive the same parameters.

\paragraph{Robot-based correction.}
FK gives the gripper's position independently of camera placement, so its predicted--observed image discrepancy can correct the extrinsics:
\begin{enumerate}[itemsep=2pt,topsep=3pt,leftmargin=*]
\item With a calibrated training camera, fit a fixed gripper-frame anchor whose projection matches the gripper-mask centroid.
\item During normal arm motion, pair its FK world position with the observed mask centroid. Keep visible, untruncated masks and the latest $60$ pairs.
\item With at least four pairs, robustly fit camera rotation/translation while fixing intrinsics. Reuse the corrected extrinsics for relational retrieval and guidance.
\end{enumerate}
The fit minimizes componentwise soft-$L_1$ reprojection error with a weak prior to the supplied pose.
Until a valid fit is available, the supplied calibration remains in use.
The depth-dependent LIBERO variant instead aligns robot-part 3D centroids with FK anchors using Kabsch fitting.

\paragraph{Sensitivity experiment.}
We perturb the extrinsics used for geometry while leaving rendered observations unchanged in ManiSkill3 OOD.
Pose error is therefore calibration error, not a rendered viewpoint change.
Object displacement is the median offline localization error.
Self-calibration starts from the incorrect parameters.

\begin{table}[ht]
\RAAGchangecolor
\centering
\caption{\textbf{Camera-extrinsic error and self-calibration.} Task success (\%). Object displacement is in cm. The exact-calibration reference is $69.6\%$; robot-part masks used for self-calibration are simulator-provided.}
\label{tab:self_calibration}
\small
\setlength{\tabcolsep}{4pt}
\begin{tabular}{@{}lrrrrr@{}}
\toprule
\makecell{Pose\\error} & \makecell{Object\\displacement} & \makecell{Frozen\\policy} &
\makecell{RAAG\\exact} & \makecell{RAAG\\incorrect} & \makecell{RAAG +\\self-calib.} \\
\midrule
$0$ & 0 & 24.0 & 69.6 & --- & 68.4 \\
$1^\circ+1$\,cm & 1.2 & 23.2 & 69.6 & 66.0 & 69.5 \\
$2^\circ+2$\,cm & 2.4 & 24.2 & 69.6 & 49.6 & 69.1 \\
$3^\circ+3$\,cm & 3.7 & 23.6 & 69.6 & 33.2 & 62.1 \\
$5^\circ+5$\,cm & 6.1 & 22.9 & 69.6 & 19.1 & 59.4 \\
\bottomrule
\end{tabular}
\end{table}

At $5^\circ+5$\,cm, self-calibration recovers $19.1\%\rightarrow59.4\%$, exceeding the frozen policy by $36.5$ points.
Deployment requires robot-part segmentation and informative visible arm motion.